\documentclass[pdflatex,sn-basic]{sn-jnl}

  \makeatletter
  \@twosidefalse
  \@mparswitchfalse
  \makeatother
  \makeatletter
\@ifundefined{@makespecialcolbox}{%
  \@ifundefined{@make@specialcolbox}{}{%
    \let\@makespecialcolbox\@make@specialcolbox
  }
}{}
\makeatother

  \usepackage{graphicx}
  \usepackage{amsmath,amssymb,amsfonts}
  \usepackage{booktabs}
  \usepackage{multirow}
  \usepackage[title]{appendix}
  \usepackage[table]{xcolor}
  \usepackage{array}
  \usepackage{paralist}
  \usepackage{tcolorbox}
  \tcbuselibrary{breakable}
  \usepackage{afterpage}
  \usepackage{hyperref}
  \usepackage{placeins}
  
  \definecolor{myboxbg}{RGB}{245,250,255}
  \definecolor{myboxframe}{RGB}{180,210,235}
  \definecolor{mytitlebg}{RGB}{180,210,235}
  \definecolor{mytitlefg}{RGB}{0,0,0}
  
  \definecolor{myblue}{RGB}{227,245,252}
  \definecolor{mygreen}{RGB}{227,253,235}
  \definecolor{imppurple}{RGB}{240,232,250}
  \definecolor{imppos}{RGB}{220,30,30}
  \definecolor{impneg}{RGB}{50,180,50}
  \newcommand{\inc}[1]{\textcolor{imppos}{\scalebox{0.82}{\,+#1}}}
  \newcommand{\dec}[1]{\textcolor{impneg}{\scalebox{0.82}{\,-#1}}}
  \newtcolorbox{theorybox}[1]{
    title={#1},
    colback=myboxbg,
    colframe=myboxframe,
    colbacktitle=mytitlebg,
    coltitle=mytitlefg,
    boxrule=0.45pt,
    arc=1.2mm,
    outer arc=1.2mm,
    fonttitle=\small\bfseries,
    fontupper=\small,
    lefttitle=2mm,
    righttitle=2mm,
    toptitle=0.7mm,
    bottomtitle=0.7mm,
    left=2mm,
    right=2mm,
    top=1.4mm,
    bottom=1.4mm,
    boxsep=0pt,
    before skip=9pt plus 2pt minus 1pt,
    after skip=8pt plus 2pt minus 1pt,
    before upper={\setlength{\parindent}{0pt}\setlength{\parskip}{0pt}}
  }

  \newtcolorbox{promptbox}[1]{
    breakable,
    title={#1},
    colback=white,
    colframe=myboxframe,
    colbacktitle=mytitlebg,
    coltitle=mytitlefg,
    boxrule=0.45pt,
    arc=1.2mm,
    outer arc=1.2mm,
    fonttitle=\footnotesize\bfseries,
    fontupper=\footnotesize,
    lefttitle=2mm,
    righttitle=2mm,
    toptitle=0.6mm,
    bottomtitle=0.6mm,
    left=2mm,
    right=2mm,
    top=1.2mm,
    bottom=1.2mm,
    boxsep=0pt,
    before skip=7pt plus 2pt minus 1pt,
    after skip=7pt plus 2pt minus 1pt,
    before upper={\setlength{\parindent}{0pt}\setlength{\parskip}{2.2pt}}
  }

  \newcommand{\blfootnote}[1]{
    \begingroup
    \renewcommand\thefootnote{}\footnote{#1}
    \addtocounter{footnote}{-1}
    \endgroup
  }
  
  \newcommand{\paratitle}[1]{\vspace{1.2ex}\noindent \textbf{#1}}


  \renewcommand{\figurename}{Figure}

  \makeatletter
  \patchcmd{\@maketitle}{\else\vskip21pt\fi}{\else\vskip17pt\fi}{}{}
  \patchcmd{\@maketitle}{\removelastskip\vskip24pt}{\removelastskip\vskip18pt}{}{}
  \patchcmd{\@maketitle}{\removelastskip\vskip24pt}{\removelastskip\vskip10pt}{}{}
  \def\keywords#1{\ifx#1\empty\else
    \def\@keywords{\par\addvspace{4pt}{\keywordfont
    {\bfseries\keywordname:} #1\par}}\fi}
  \makeatother
  
\begin{document}

  \title[MoCA]{MoCA: Implicit Social Context Analysis}

  \author[1]{\fnm{Wenhao} \sur{Xu}}
  \email{e1561245@u.nus.edu}
  \equalcont{Equal contributors.}

  \author[1]{\fnm{Kaiwen} \sur{Zhang}}
  \email{kaiwen\_zhang24@u.nus.edu}
  \equalcont{Equal contributors.}
  \author[1,3]{\fnm{Hao} \sur{Li}}\email{sc.lihao@whu.edu.cn}
  \author[1]{\fnm{Maowei} \sur{You}}\email{youmaowei@u.nus.edu}
  \author[1]{\fnm{Yongzheng} \sur{Ji}}\email{yongzheng.ji@u.nus.edu}
  \author[1]{\fnm{Siyuan} \sur{Zuo}}\email{e1561374@u.nus.edu}
  \author[1]{\fnm{Jingxuan} \sur{Yu}}\email{e1553436@u.nus.edu}
  \author[1]{\fnm{Sina} \sur{A}}\email{sina001@e.ntu.edu.sg}
  \author[1]{\fnm{Xinyao} \sur{Tan}}\email{e1519782@u.nus.edu}
  \author[1]{\fnm{Bobo} \sur{Li}}\email{libobo@nus.edu.sg}
  \author*[2]{\fnm{Hao} \sur{Fei}}\email{haofei7419@gmail.com}
  \author[1]{\fnm{Mong-Li} \sur{Lee}}\email{leeml@comp.nus.edu.sg}
  \author[1]{\fnm{Wynne} \sur{Hsu}}\email{whsu@comp.nus.edu.sg}

  \affil[1]{\orgname{National University of Singapore},
  \orgaddress{\city{Singapore}, \country{Singapore}}}
  
  \affil[2]{\orgname{University of Oxford},
  \orgaddress{\city{Oxford}, \country{United Kingdom}}}
  
  \affil[3]{\orgname{Wuhan University},
  \orgaddress{\city{Wuhan}, \country{China}}}

  \abstract{

  \enlargethispage{3\baselineskip}
  Human social communication (e.g., affection, intent) is often conveyed in highly implicit ways, where underlying meanings are expressed through indirect, socially and culturally grounded signals rather than explicit statements.
  Such implicit social contexts are pervasive in real-world interactions, yet unfortunately there is still a lack of a formal and systematic framework to study them.
  In this paper, we introduce \textit{Implicit Social Context Analysis} (MoCA), a novel task that systematically models implicit social scenarios along three key dimensions: affection, intent, and stance.
  We construct a high-quality benchmark of 3,108 multimodal instances collected from real-world sources, with fine-grained cognitive annotations to reveal who expresses what toward whom, and how and why it is conveyed.
  With MoCA data, we show that state-of-the-art multimodal large language models struggle significantly on MoCA, due to their reliance on explicit cues and limited capability in reasoning over latent social context.
  To address this challenge, we propose a novel \textit{Conflict-Driven Abductive Reasoning} (CoDAR) framework, which models the discrepancy between observed expressions and expected truthful behavior as cognitive conflict, enabling the inference of hidden mental states.
  Extensive experiments demonstrate that CoDAR substantially improves model performance, while a large gap to human reasoning remains, highlighting the fundamental difficulty of implicit social understanding.

  }

  \afterpage{
    \blfootnote{All our data and code (\href{https://github.com/Coder12188/MoCA}{Coder12188/MoCA}) will be made open to facilitate future research.}
  }
  \maketitle
  \clearpage

  \section{Introduction}
  
  Human social interaction has evolved into a highly complex system, where communication and expression are deeply shaped by cultural background, social roles, and contextual dependencies~\citep{grice1975logic, goffman2023presentation}.
  Constrained by power structures, politeness norms, and self-protection considerations, people in everyday interactions often avoid expressing their true emotions, thoughts, or stances in a straightforward manner~\citep{brown1987politeness}.
  Instead, they tend to strategically rely on more indirect and implicit forms of expression, cognitively choosing ways to convey underlying meanings without stating them explicitly~\citep{searle1975indirect}.
  In both offline interactions and online social platforms, such expressions are frequently realized through subtle pragmatic cues, facial micro-expressions, vocal tone variations, fine-grained body movements, and creative multimodal artifacts such as memes or short videos, often involving irony, sarcasm, metaphor, or indirect insinuation to express complex emotions and intentions~\citep{van2020hierarchy,van2016social,cheshin2020impact}.
  Despite its prevalence and importance, this phenomenon remains largely underexplored, and there has been scarce systematic effort to study it in a unified manner.
  Existing work either focuses on implicit affect and intent analysis in purely textual settings~\citep{fei2023reasoning,villarroel2017unveiling,alswaidan2020survey}, whereas real-world implicit social expression is inherently multimodal; or considers multimodal tasks such as sarcasm detection~\citep{saha2025mustreason}, which are typically limited to shallow label classification without requiring deeper cognitive-level interpretation of underlying implications; or relies on highly abstract theory-of-mind frameworks~\citep{kanske2018social} that are often detached from concrete implicit social scenarios.
  To bridge this gap, this paper is dedicated to a systematic investigation of this form of implicit social computation.

  \begin{figure}[!t]
    \centering
    \includegraphics[width=\textwidth]{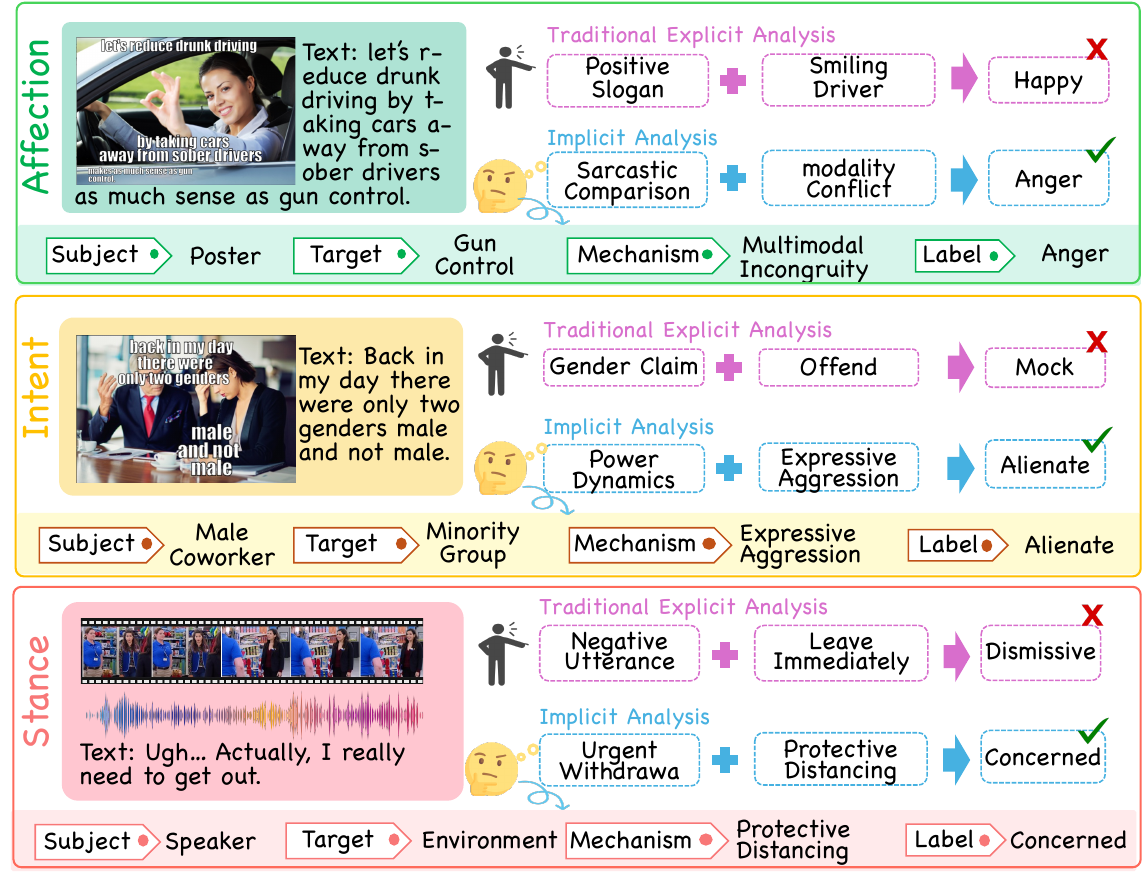}
    \caption{
    Examples of real-world implicit social expressions, where the underlying real affection, intent, or stance must be analyzed implicitly from the given multimodal cues.
    }
    \label{fig:intro}
  \end{figure}

  In this work, we formalize these scenarios as \textit{Implicit Social Context Analysis} (\textbf{MoCA}).
  Recent benchmarks motivate a structured treatment of latent social meaning rather than a single generic label.
  Multimodal affective benchmarks combine linguistic, visual, and acoustic evidence for emotion reasoning~\citep{cheng2024emotionllama}, while visually grounded dialogue benchmarks examine affective reasoning in conversational context~\citep{haydarov2024affectivevisualdialog}.
  Multimodal Theory-of-Mind and intent benchmarks infer goals, beliefs, plans, and communicative purposes from situated behavior~\citep{jin2024mmtom,zhang2024mintrec2}.
  Multimodal stance detection further treats position as target-dependent and jointly grounded in textual and visual evidence~\citep{liang2024multimodalstance}.

  MoCA therefore distinguishes three related but non-equivalent inference targets: an underlying affective state, an intended outcome, and a target-relative position:
  \begin{compactitem}
      \item \textbf{\textit{Implicit Affection}} infers a concealed or indirectly communicated affective state from ambiguous or conflicting verbal content, prosody, facial behavior, and visual context. It targets the underlying emotion rather than the overt affective display.
      \item \textbf{\textit{Implicit Intent}} infers the outcome sought through an observed expression or action. Although the behavior may be explicit, its motivating goal must be inferred from the addressee, interaction history, and plausible social consequences.
      \item \textbf{\textit{Implicit Stance}} infers the subject's evaluative or relational position toward a person, group, topic, or proposition. It covers support, opposition, distancing, alignment, and power-sensitive positioning grounded in multimodal context. Because stance is relational, its target is inferred rather than treated as metadata.
  \end{compactitem}
  These dimensions may co-occur but remain distinct: affection represents an affective state, intent an intended outcome, and stance a target-relative relation.
  Each instance is assigned to a primary scenario that constrains its label and mechanism spaces; Figure~\ref{fig:intro} provides representative examples.

  To facilitate research along this direction, we construct a large-scale benchmark dataset for MoCA.
  We first collect data from a wide range of real-world scenarios, including memes, online discussions, public debates, and situational comedies, all of which contain rich multimodal signals such as text, images, and videos, from which we categorize instances into affection, intent, and stance scenarios.
  For each instance, we further annotate a comprehensive set of labels from a human cognitive perspective, including who is expressing (\textbf{subject}), toward whom the expression is directed (\textbf{target}), what the underlying meaning is (\textbf{label}), and how\&why the meaning is conveyed (\textbf{mechanism}).
  We design a human annotation pipeline with rigorous verification procedures to ensure high-quality control, resulting in a MoCA dataset of 3,108 instances.

  The MoCA task is inherently non-trivial.
  Although state-of-the-art (SoTA) multimodal large language models (MLLMs)~\citep{gpt5systemcard2025, gemini2025} have shown strong performance across many cross-modal tasks, they consistently struggle when applied to MoCA scenarios.
  The key reason is that many implications in implicit social contexts are not explicitly present in textual or visual cues, whereas MLLMs are primarily designed to interpret and reason over observable signals.
  We argue that understanding such implicit meanings requires access to latent social context, such as interpersonal relationships, cultural norms, historical interactions, and power structures, as well as human-like cognitive reasoning built upon these factors to infer the subject's hidden mental states.
  To address this challenge, we propose a novel \textit{Conflict-Driven Abductive Reasoning} (\textbf{CoDAR}) framework.
  Grounded in cognitive-driven insights, we observe that in implicit social scenarios, individuals often adopt expressions that deviate from what would be expected if their true internal states were directly revealed, due to underlying goals, constraints, and perceived risks.
  This suggests that the core of MoCA reasoning lies in identifying and interpreting the \textbf{cognitive ``conflict''} embedded in the context, and CoDAR explicitly models the discrepancy between the subject's observed expression and their expected expression under truthful conditions, leveraging abductive reasoning to recover the underlying affection, intent, and stance.

  We conduct extensive experiments on the MoCA dataset with both general-purpose multimodal models and domain-adapted variants, covering both open-source and proprietary SoTA MLLMs, and observe a substantial performance gap between these models and human-level reasoning in implicit social scenarios, highlighting the inherent challenge of the task.
  Notably, equipping these models with our proposed CoDAR framework leads to significant performance improvements.
  We further provide in-depth analyses of different models and approaches on MoCA, examining their failure modes and exploring potential directions for improvement.
  Overall, our main contributions can be summarized as follows.
  \begin{compactitem}
      \item We introduce a novel task of Implicit Social Context Analysis, providing the first systematic study of how AI can perform human-level cognitive reasoning in implicit social scenarios.
      \item We construct a high-quality, real-world multimodal dataset of 3,108 MoCA instances spanning affection, intent, and stance, and extensively evaluate across SoTA models.
      \item We propose a cognition-aware conflict-driven reasoning framework, CoDAR, as a benchmark method for implicit social inference, which effectively enhances the performance of current multimodal large models.
  \end{compactitem}

  \section{Related Work}
  
  \subsection{Implicit Social Computing and Social Cognition Analysis}
  
  Social computing seeks to endow machines with socially aware capabilities through AI algorithms, enabling computational systems to perceive, model, and respond to human social signals.
  It has emerged as an important research direction with a wide range of real-world applications~\citep{wang2021survey,wu2021modeling}.
  Within this broad agenda, analyzing implicit social scenarios is particularly valuable because it more closely reflects how social interactions unfold in practice~\citep{zhou2021implicit}.
  Unlike explicit settings, where the relevant social state is directly stated or readily observable, implicit analysis requires recovering unstated affect, intention, stance, and communicative goals from contextual evidence.
  This requirement shifts the problem from surface-level recognition toward structured inference over latent social variables and the observable cues through which they are indirectly expressed.
  
  In the NLP domain, several lines of work have explored individual aspects of implicit social understanding.
  For example, implicit sentiment analysis~\citep{wei2020bilstm} seeks to recognize emotions and affective attributes that are not directly expressed.
  Instead, such states may be conveyed through sarcasm, passive aggressiveness, or multimodal incongruity~\citep{wu2021modeling}.
  Similarly, intent understanding~\citep{louvan2020recent} aims to infer user intentions in task-oriented settings~\citep{zhang2016joint}.
  A representative example is identifying destination or time preferences in flight-booking scenarios~\citep{jbene2025intent}.
  These formulations provide useful supervision for recovering specific latent variables, but they commonly prioritize predefined labels, slots, or task-completion goals.
  Consequently, they do not necessarily model why an utterance adopts an indirect form, toward whom it is directed, or which social mechanism produces its implied meaning.
  
  However, these existing studies are largely confined to unimodal language settings.
  In contrast, real-world implicit social interactions are inherently multimodal.
  Large-scale resources such as CMU-MOSEI and MELD demonstrate that linguistic, visual, and acoustic streams provide complementary evidence for affective understanding in naturalistic interactions~\citep{zadeh2018cmumosei,poria2019meld}.
  Individuals frequently combine linguistic signals with visual cues, including facial expressions and meme imagery, to convey conflicting or indirect meanings and achieve subtle social goals~\citep{farabi2024survey}.
  Such signals are not always redundant: agreement can reinforce an interpretation, whereas cross-modal conflict can reverse the literal meaning or expose a concealed communicative intention.
  Multimodal sarcasm detection~\citep{castro2019towards,cai2019multimodal,qin2023mmsd2,guo2025multi} is perhaps the most closely related line of work.
  Nevertheless, its task formulation remains relatively shallow, typically focusing on label classification without requiring deeper reasoning about the underlying implications.
  In particular, predicting a sarcasm label alone does not require a model to identify the social actors, recover their relational roles, explain the operative mechanism, or verify the resulting interpretation.
  
  Another related line of research focuses on abstract social cognition modeling.
  Social-IQ frames artificial social intelligence as question answering over in-the-wild videos, requiring models to reason about interactions rather than isolated affective labels~\citep{zadeh2019socialiq}.
  Recent work on multimodal Theory of Mind~\citep{jin2024mmtom, villa2025moments, shi2025muma} and explanatory reasoning has moved beyond explicit grounding toward richer forms of socially informed inference.
  Representative efforts include \textit{MuMA-ToM}, \textit{MoMentS}, \textit{Multimodal UNcommonsense}, and \textit{DixitWorld}~\citep{shi2025muma, villa2025moments, son2026multimodal, mo2025dixitworld}.
  Collectively, these studies examine mental-state inference, abductive hypothesis generation, explanatory reasoning, and the interpretation of events that cannot be resolved through direct perception alone.
  Complementary work on abductive inference and commonsense knowledge generation recovers plausible latent causes, intentions, and surrounding events from partial observations~\citep{bhagavatula2020abductive,park2020visualcomet,bosselut2019comet}.
  Nevertheless, these settings often simplify the social space into low-dimensional mental variables, constrained game-like environments, or physically unusual events.
  These abstractions are useful for isolating particular cognitive capabilities, but they do not fully capture the deeply masked and strategically constructed expressions that characterize natural implicit interactions.
  In such interactions, literal content, multimodal evidence, interpersonal relationships, and socio-cultural expectations may jointly determine the intended meaning.
  To address these limitations, this work presents, to our knowledge, the first systematic study and benchmark of implicit social context analysis in a multimodal setting.
  
  \subsection{Multimodal Foundation Models for Social Reasoning}
  In the era of small-scale neural networks, AI systems were largely incapable of handling or reasoning about such subtle and complex social phenomena~\citep{sap2019social, vinciarelli2009social, forbes2020social, mathur2024advancing}.
  These systems were generally optimized for narrow prediction objectives and local correlations, with limited access to the broad knowledge needed for context-sensitive social inference.
  Recent advances in multimodal large language models (MLLMs) have demonstrated strong potential and substantially improved reasoning over multimodal inputs.
  This progress is especially evident when models use multimodal Chain-of-Thought and compositional reasoning techniques~\citep{gao2025interleaved, mitra2024compositional, wei2022chain, zhang2023multimodal, fei2023reasoning, fei2024videoofthought, zhang2025improve, li2025vegas}.
  These techniques can decompose a complex query into intermediate subproblems, organize heterogeneous evidence, and make parts of the inference process more explicit.
  
  Despite this progress, most existing approaches remain limited to relatively shallow forms of social reasoning that rely on explicit cues in the input~\citep{mathur2025social}.
  Their evaluations also frequently emphasize answer or label correctness, providing limited evidence that the inferred social explanation is faithful to the multimodal observations.
  Recent probing likewise shows that MLLMs can overlook implicit inconsistencies even when the relevant perceptual and reasoning capabilities are available~\citep{yan2025hidden}.
  In complex implicit social scenarios, these models often fail because successful reasoning requires cognitive-level understanding rather than direct cue matching.
  Required knowledge includes social relationships, cultural norms, and background context, together with the ability to uncover deeper sources of conflict~\citep{ziems2023normbank}.
  The model must explain why a particular expression is chosen, what social motivations underlie it, and how apparently inconsistent signals support a coherent interpretation.
  The difficulty therefore extends beyond combining modalities: a reasoner must identify the expected social pattern, locate deviations from that expectation, and infer a plausible latent mechanism.
  It must also jointly recover the subject, target, mechanism, and latent social state, while checking whether these components remain mutually consistent.
  Therefore, this work proposes an effective solution by introducing a Conflict-Driven Abductive Reasoning framework.
  The framework treats conflict as an informative reasoning signal and progressively connects explicit perception, social context, expectation modeling, abductive inference, and consistency verification.

  \begin{figure}[!t]
  \centering
  \includegraphics[width=\textwidth]{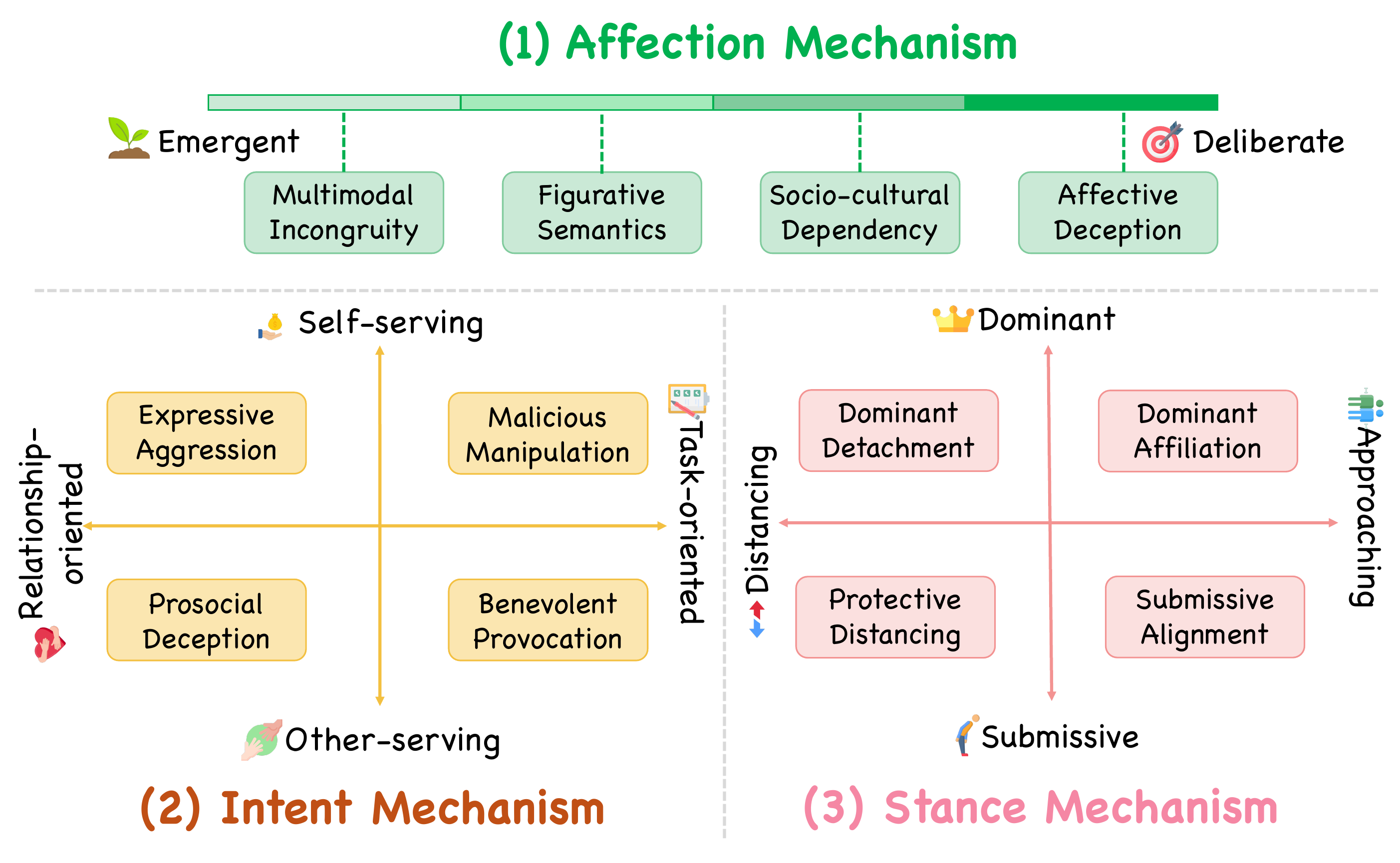}
  \caption{The mechanism spaces of implicit social contexts.}
  \label{fig:implicit}
  \end{figure}

  \section{Approaching Implicit Social Context Analysis}

  \subsection{Task Definition}
  Here we try to lay a formal definition of the MoCA task.
  MoCA recovers a structured implicit social interpretation, rather than a single label, from the multimodal input shown in Figure~\ref{fig:intro}.
  Let $x_{\text{txt}}$, $x_{\text{img}}$, $x_{\text{vid}}$, and $x_{\text{aud}}$ denote the textual, image, video, and audio inputs.
  For scenario $s \in \{\text{Affection}, \text{Intent}, \text{Stance}\}$, the task is
  \begin{equation}
  (x_{\text{txt}}, x_{\text{img}}, x_{\text{vid}}, x_{\text{aud}} ) \mapsto (y_{\text{subj}}, y_{\text{tgt}}, y^s_{\text{mech}}, y^s_{\text{label}}) \,,
  \end{equation}
  where $y_{\text{subj}}$ and $y_{\text{tgt}}$ identify the expressing subject and intended target, and $y^s_{\text{mech}}$ and $y^s_{\text{label}}$ specify the realizing mechanism and latent label.
  The superscript $s$ indicates that mechanism and label are selected from scenario-specific spaces.

  The three scenarios specify different inference targets.
  \textit{Affection} covers the broader domain of latent affective states, not only interpersonal fondness, whose experience and observable expression may diverge~\citep{scherer2005emotion,gross1998emotionregulation}.
  \textit{Intent} covers goal-directed states that organize action through beliefs, desires, and anticipated behavior~\citep{ajzen1991plannedbehavior,malle1997intentionality}.
  \textit{Stance} covers evaluative and relational positioning toward a target~\citep{dubois2007stance}.
  These domains may co-occur; $s$ marks the primary inference target and does not imply psychological independence.

  \subsection{Mechanism Space}
  MoCA does not assume that a latent label can be recovered from one modality or simple cue aggregation.
  It also predicts the \textit{\textbf{mechanism}} through which the observed expression constructs or conceals that label.
  Each scenario contains four mechanisms:
  \begin{compactitem}
      \item In Affection, mechanisms range from cross-modal semantic conflict (\textit{Multimodal Incongruity}) to deliberate emotional masking (\textit{Affective Deception});
      \item In Intent, mechanisms range from overt but strategically framed hostility (\textit{Expressive Aggression}) to covert behavioral manipulation (\textit{Malicious Manipulation});
      \item In Stance, mechanisms range from power-asymmetric disengagement (\textit{Dominant Detachment}) to strategically compliant positioning (\textit{Submissive Alignment}).
  \end{compactitem}
  Figure~\ref{fig:implicit} presents the complete mechanism spaces.
  A valid mechanism must explain the predicted label given the observed expression and recovered subject--target relation.

  \subsection{Evaluation}
  \label{sec:evaluation}
  Evaluation is scenario-wise: models and humans are scored on the same $N_s$ instances.
  Each response is parsed into $(\mathrm{Subject},\mathrm{Target},\mathrm{Mechanism},\mathrm{Label})$ and compared with its annotation by normalized exact match.
  We lowercase strings, map hyphens and underscores to spaces, and collapse repeated separators; missing, unparsable, or scenario-invalid fields are incorrect, without dropping the sample.
  For $\mathcal K=\{\mathrm{subj},\mathrm{tgt},\mathrm{mech},\mathrm{label}\}$, let $c_{i,k}$ indicate a normalized match on field $k$.
  \textbf{Atomic Component Accuracy} reports $\mathrm{Acc}_k=N_s^{-1}\sum_i c_{i,k}$ for each field;
  \textbf{Dyadic Alignment} is $\mathrm{Dyad}=N_s^{-1}\sum_i c_{i,\mathrm{subj}}c_{i,\mathrm{tgt}}$;
  \textbf{Mechanism Faithfulness} is $\mathrm{Faith}=\sum_i c_{i,\mathrm{label}}c_{i,\mathrm{mech}}/\sum_i c_{i,\mathrm{label}}$; and
  \textbf{Holistic Alignment} is $\mathrm{Holis}=N_s^{-1}\sum_i\prod_{k\in\mathcal K}c_{i,k}$.
  These metrics test component recovery, directed relations, output-level mechanism--label consistency, and complete structured prediction, respectively.
  Faithfulness does not claim access to the model's internal reasoning.
  Scores are reported as percentages.

  \section{Benchmarking Implicit Social Context Analysis}

  \subsection{MoCA Dataset Construction}
  
  Unlike conventional datasets focusing on sentiment polarity or semantic consistency, MoCA captures implicit social semantics via structured relational tuples.
  Our construction pipeline entails four stages as follows.

  \paratitle{Data Sources and Preprocessing}.
  To cover diverse implicit social expressions, we merge candidate samples from two complementary sources into a unified pool for downstream filtering and annotation. The first comprises existing multimodal datasets: MUStARD~\citep{castro2019towards} and MIntRec~\citep{zhang2022mintrec} for video, and RedCaps~\citep{desai2021redcaps}, MMSD2.0~\citep{qin2023mmsd2}, Hateful Memes~\citep{kiela2020hatefulmemes}, MultiMET~\citep{zhang2021multimet}, MemeCap~\citep{hwang2023memecap}, and additional Instagram captions for image-text data. The second consists of supplementary collected data, such as meme-text pairs and online debates, yielding rich indirect expressions shaped by semantic contrast, socio-cultural contexts, and pragmatics.

  \paratitle{Unimodal Consistency Filtering}. To eliminate modality-specific biases and ensure genuine cross-modal dependency, we implement a consistency filtering step. For each candidate, we evaluate predictions under three settings:visual-only ($R_{\text{img}}, R_{\text{vid}}$), text-only ($R_{\text{txt}}$), and multimodal ($R_{\text{multi}}$). A sample is retained only if the multimodal interpretation diverges from both unimodal counterparts:
  \begin{equation}
  R_{\text{multi}} \notin \{ R_{\text{img}}, R_{\text{vid}}, R_{\text{txt}} \}
  \end{equation}
  This strategy filters out instances inferable from a single modality, ensuring the dataset necessitates holistic multimodal reasoning.

  \paratitle{Cross-Modal Structured Annotation}. 
  Filtered samples are annotated via a 7-tuple structured schema: (Scenario, Domain, Culture, Subject, Target, Mechanism, Label). 
  Instead of isolated categorical labels, this schema deconstructs implicit social meaning into relational units. 
  Specifically, \textit{Mechanism} encodes how implicit meaning is realized (e.g., cross-modal conflict, figurative expression, or socio-cultural dependency), facilitating evaluation of both predictive outcomes and reasoning structures. \textit{Subject} and \textit{Target} transcend visually explicit entities to include implicit or socially presupposed individuals and groups.

  \paratitle{Quality Control and Final Screening}. 
  Because implicit social interpretation is inherently subjective and context-dependent, we employ a multi-stage quality control protocol to ensure annotation consistency and logical validity. The process comprises independent annotation, cross-checking, and arbitration. Samples are first annotated under unified guidelines, followed by consistency verification to identify conflicts, and finally adjudicated by experienced annotators to determine whether disputed cases should be revised, retained, or discarded. All retained samples undergo final validation to ensure coherent scenario assignment, valid Subject--Target relations, pragmatically plausible Mechanisms, and interpretations supported by multimodal evidence.

  \subsection{Dataset Statistics and Highlights}
  
\begin{figure}[!t]
  \centering
  \includegraphics[width=\textwidth]{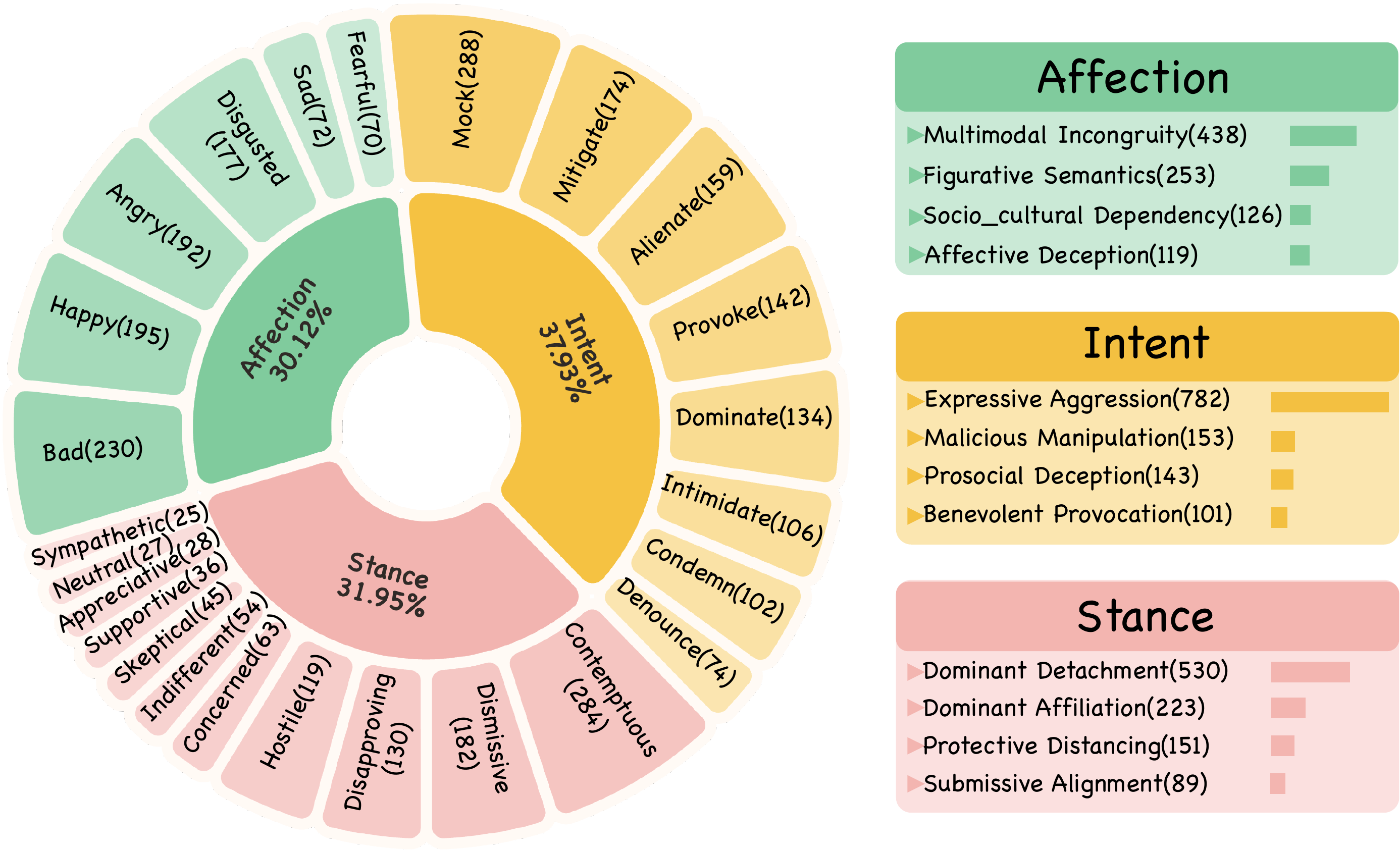}
  \caption{
  Distributions of labels and scenarios (left) and implicit-social mechanisms (right).
  }
  \label{fig:label_dist}
  \end{figure}

  MoCA spans diverse social interaction contexts, from public discourse to interpersonal communication, covering both image-text and video-text modalities across varied socio-cultural settings. After multi-stage filtering and manual verification, the dataset contains 3,108 high-quality multimodal samples. Dataset statistics are summarized in Table~\ref{tab:moca_stats_split_line} and Figure~\ref{fig:label_dist}.
  
    \begin{table}[!t]
    \centering
    \caption{General statistics of MoCA dataset.}
    \label{tab:moca_stats_split_line}
    \small
    \setlength{\tabcolsep}{4pt}
    \renewcommand{\arraystretch}{1.05}
  
    \begin{tabular*}{\textwidth}{@{\extracolsep{\fill}} l ccccccc @{}}
      \toprule
  
      \rowcolor[gray]{.9}
      \multicolumn{8}{c}{\textbf{Domain}} \\
      \midrule
  
      & Networked & Occupational & Civic & Intimate & Familial & Peer & Educational \\
  
      \multirow{2}{*}{\textbf{Num.}}
      & 1268 & 532 & 440 & 411 & 220 & 176 & 61 \\
      & (40.80\%) & (17.12\%) & (14.16\%) & (13.22\%) & (7.08\%) & (5.66\%) & (1.96\%) \\
  
      \midrule

      \rowcolor[gray]{.9}
      \multicolumn{6}{c}{\textbf{Culture}}
      & \multicolumn{2}{c}{\textbf{Modality}} \\
  
      \cmidrule{1-6} \cmidrule{7-8}
  
      & General & M. East & N. Am. & S. Asian & E. Asian & I+T & V+T \\
  
      \multirow{2}{*}{\textbf{Num.}}
      & 2884 & 123 & 70 & 17 & 14 & 2230 & 878 \\
      & (92.79\%) & (3.96\%) & (2.25\%) & (0.55\%) & (0.45\%) & (71.75\%) & (28.25\%) \\
  
      \bottomrule
  
    \end{tabular*}
  
  \end{table}
  
  \paratitle{Implicit Social Scenarios and Labels}. Figure~\ref{fig:label_dist} shows the distribution of three implicit social scenarios: Affection (30.12\%), Intent (37.93\%) and Stance (31.95\%). The inner ring denotes scenario categories, and the outer ring the fine-grained label composition within each scenario. Affection contains emotion-related labels (happy, bad, angry, disgusted, sad, fearful).  Intent contains social intention labels (mock, mitigate, alienate, provoke, dominate, intimidate, condemn, denounce). Stance contains stance-related labels (contemptuous, dismissive, disapproving, hostile, concerned, indifferent, skeptical, supportive, sympathetic, neutral, appreciative). This hierarchical structure supports modeling at both scenario and fine-grained social semantic levels.
  
  \paratitle{Diverse Domains in Social Interaction Contexts}. Table~\ref{tab:moca_stats_split_line} shows coverage across multiple social domains, including Networked, Occupational, Civic, Intimate, Familial, Peer, and Educational contexts. The benchmark spans both public and interpersonal environments, enabling evaluation of implicit reasoning across heterogeneous communicative settings and social relations.
  
  \paratitle{Cultural Diversity and Contextual Variation}. Table~\ref{tab:moca_stats_split_line} reports the distribution of socio-cultural contexts. Most samples belong to General Context; the remainder span Middle Eastern, North American, South Asian, and East Asian contexts. Models therefore require background knowledge to interpret this variation.

  \paratitle{Complementary Multimodal Coverage}. MoCA includes image-text and video-text inputs. Image-text samples provide compact symbolic context, often through cross-modal semantic contrast, while video samples add temporal cues such as tone variation, actions, and interaction dynamics. Together, these static and dynamic inputs broaden the evidence available for multimodal implicit reasoning.

  \paratitle{Implicit Mechanism Characteristics}. MoCA explicitly annotates Mechanism in addition to Label to characterize how implicit meaning is constructed. Across the dataset, implicit meaning is realized through mechanisms including cross-modal semantic conflict, figurative expression, socio-cultural dependency, affective masking, and socially strategic expression. Explicit modeling of Mechanism makes the reasoning process analyzable and enables evaluation of whether models capture the formation of implicit meaning.

  \section{CoDAR: Conflict-Driven Abductive Reasoning}
  \subsection{Theoretical Motivation}
  
  The MoCA task requires models to jointly recover structured, latent social information from multimodal inputs. Unlike tasks that map observable cues directly to labels, MoCA must explain why an observed expression takes a specific form in a given situation. We ground this requirement in three complementary frameworks from cognitive and social psychology.
  
  \begin{theorybox}{Impression Management}
  Impression management theory argues that overt social behavior is shaped by situational pressures, anticipated audiences, and interactional goals~\citep{sezer2022impression}. Consequently, observable expressions do not always provide transparent access to a person's underlying affection, intent, or stance. Individuals may instead suppress, reframe, or strategically mask internal states to protect relationships, preserve status, avoid sanctions, or pursue context-dependent social outcomes. In MoCA, explicit expression $x$ is therefore socially produced evidence, not a direct readout of the latent state.
  \end{theorybox}
  
  \begin{theorybox}{Expectancy Violations Theory}
  Expectancy Violations Theory proposes that people continuously form context-sensitive expectations about how others should behave within a given situation~\citep{burgoon2015expectancy,buidze2025expectation}. We denote this normative expectation by $e$, which is conditioned on social roles, cultural norms, event structure, and relational dynamics. When the observed expression $x$ departs from $e$, the deviation becomes salient because it violates the behavior predicted by the context. Importantly, this conflict is not any superficial mismatch between modalities; it is a socially meaningful discrepancy between observed conduct and contextual expectation. We denote this discrepancy by $\mathcal{C}$, thereby converting implicit understanding into the problem of identifying and explaining a context-grounded violation.
  \end{theorybox}
  
  \begin{theorybox}{Theory of Mind}
  Theory of Mind describes the human capacity to attribute beliefs, motives, intentions, and other latent mental states to social actors~\citep{lake2017building,byom2013theory}. After a conflict is detected, an observer must infer which hidden state or strategic objective would make the otherwise unexpected expression rational. This inference is abductive because the underlying cause is not directly observed and must be recovered as the explanation that best accounts for available evidence. For MoCA, this perspective connects conflict resolution to the structured recovery of the subject, target, communicative mechanism, and latent social label.
  \end{theorybox}
  
  Together, these theories define a coherent account of implicit social reasoning. Impression management explains why overt behavior can mask internal states; expectancy violations identify where that masking produces a salient discrepancy; and Theory of Mind explains how interpreters recover its cause. Let $x$ denote the observed explicit expression and $e$ the normative expectation under the current social context. We formalize the semantic discrepancy between them as a conflict $\mathcal{C}$. Consequently, recovering latent social meaning is fundamentally equivalent to constructing an abductive explanation of the root cause of $\mathcal{C}$.

  \subsection{Reasoning Architecture}
  
\begin{figure}[!t]
    \centering
    \setlength{\abovecaptionskip}{0pt}
    \includegraphics[width=0.865\linewidth]{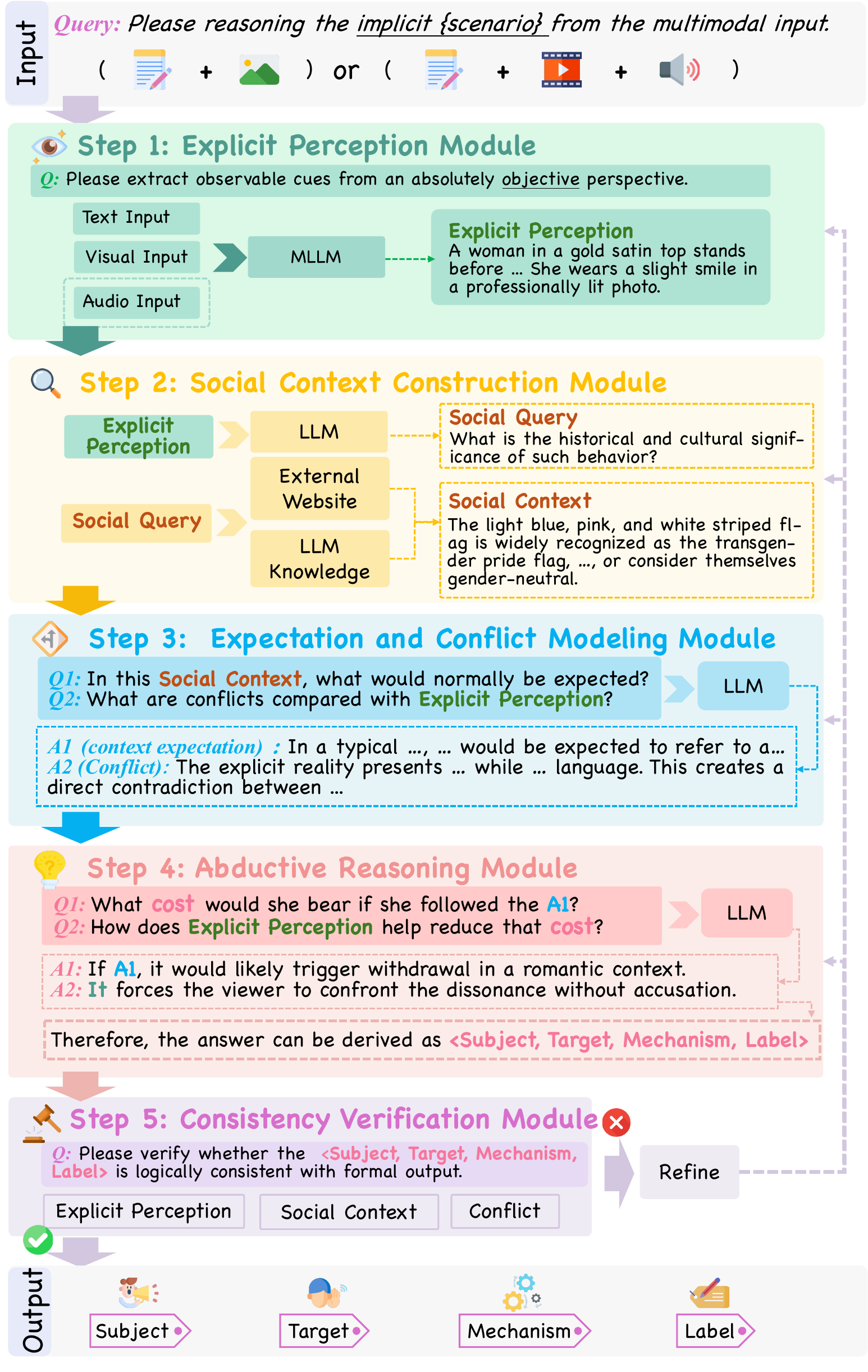}
    \caption{Overview of CoDAR, a conflict-driven abductive reasoning chain for recovering structured implicit social meaning from multimodal input.}
    \label{fig:methodology}
    \vspace{-13pt}
  \end{figure}

  Building on the preceding theoretical motivation, we introduce Conflict-Driven Abductive Reasoning (CoDAR), a five-step framework for structured implicit social understanding. CoDAR treats the discrepancy between observable evidence and contextual expectation as the central object of reasoning. As illustrated in Figure~\ref{fig:methodology}, the framework supports both image-text and video-text inputs. Its five reasoning steps are organized into three functional stages that connect multimodal evidence, social context, expectation conflict, and structured interpretation.
  
  \paratitle{Stage I: Multimodal Evidence Grounding.}
  The first stage corresponds to the Explicit Perception Module (Step~1). It extracts directly observable cues from the multimodal input while avoiding premature interpretation of latent social meaning. In the image-text setting, the module records textual content together with static visual cues. In the video-text setting, it additionally captures temporally extended evidence. Such evidence includes scene evolution, bodily behavior, interaction dynamics, and other event-level cues unfolding over time. Speech contained in a video is converted into textual transcripts and processed jointly with the visual stream. These observations are consolidated into an explicit perception representation, denoted by $x$, which provides the evidentiary basis for subsequent reasoning.
  
  \paratitle{Stage II: Contextual Expectation and Conflict Construction.}
  The second stage combines the Social Context Construction Module (Step~2) with the Expectation and Conflict Modeling Module (Step~3). Given $x$, Step~2 first identifies the background knowledge required to interpret the observed expression. It then formulates focused questions concerning relevant facts, event connections, and social norms. Each query can be assigned to a model-prior, external web-search, or hybrid retrieval route. The resulting evidence is synthesized into three complementary fields: \textit{fact}, \textit{connection}, and \textit{social norm}. Together, these fields form a social context representation $c$ that situates the observation within its broader social environment.
  This routing follows the broader principle of retrieval-augmented and reasoning-action systems, in which external evidence supplements parametric knowledge when the observation alone is insufficient~\citep{lewis2020rag,yao2023react}.
  
  Based on $c$, Step~3 constructs a context-conditioned expectation $e$. This expectation describes what would normally occur under the given situation according to the relevant norms, event structure, and relational dynamics. CoDAR then compares the explicit perception $x$ with $e$ and extracts their socially meaningful discrepancy:
  \begin{equation}
      \mathcal{C} = \Delta(x,e).
  \end{equation}
  Here, $\mathcal{C}$ identifies which observable cue departs from which contextual expectation. The module also formulates an abductive question asking why the explicit reality is presented despite that expectation. This transformation shifts the task from directly assigning a latent label to explaining why the context-grounded deviation arises.
  
  \paratitle{Stage III: Staged Abductive Inference and Verification.}
  The final stage integrates the Abductive Reasoning Module (Step~4) with the Consistency Verification Module (Step~5). Given $\mathcal{C}$, Step~4 performs a prompt-guided sequence of counterfactual analysis, strategic interpretation, participant recovery, and taxonomy-guided output selection. It produces a structured interpretation
  \begin{equation}
      \mathcal{A}
      =
      \{y_{\text{subj}},y_{\text{tgt}},
      y^s_{\text{mech}},y^s_{\text{label}}\},
  \end{equation}
  where the subject and target are selected from instance-specific candidates. The mechanism and label are selected from the output spaces defined for scenario $s$.
  
  Section~\ref{sec:conflict_constrained_inference} describes this staged inference process in detail. Step~5 evaluates the recovered interpretation through a Boolean verification function:
  \begin{equation}
      \Phi(\mathcal{A},x,c,\mathcal{C}) \in \{0,1\}.
  \end{equation}
  The verifier checks whether $\mathcal{A}$ is supported by the explicit evidence, compatible with the social context, and coherent with the identified conflict. When all checks pass, $\Phi=1$, and the structured interpretation is accepted. Otherwise, $\Phi=0$, and the verifier returns a rejection status together with a description of the detected breakpoint. This feedback initiates a revision round that reconstructs the social context, conflict, and abductive interpretation before verification is performed again. The cycle terminates when the interpretation is accepted or the predefined revision budget is exhausted.

  \vspace{0.5\baselineskip}
  \subsection{Conflict-Constrained Staged Abductive Inference}
  \label{sec:conflict_constrained_inference}

  \begin{figure}[!t]
    \centering
    \setlength{\abovecaptionskip}{0pt}
    \includegraphics[width=0.99\linewidth]{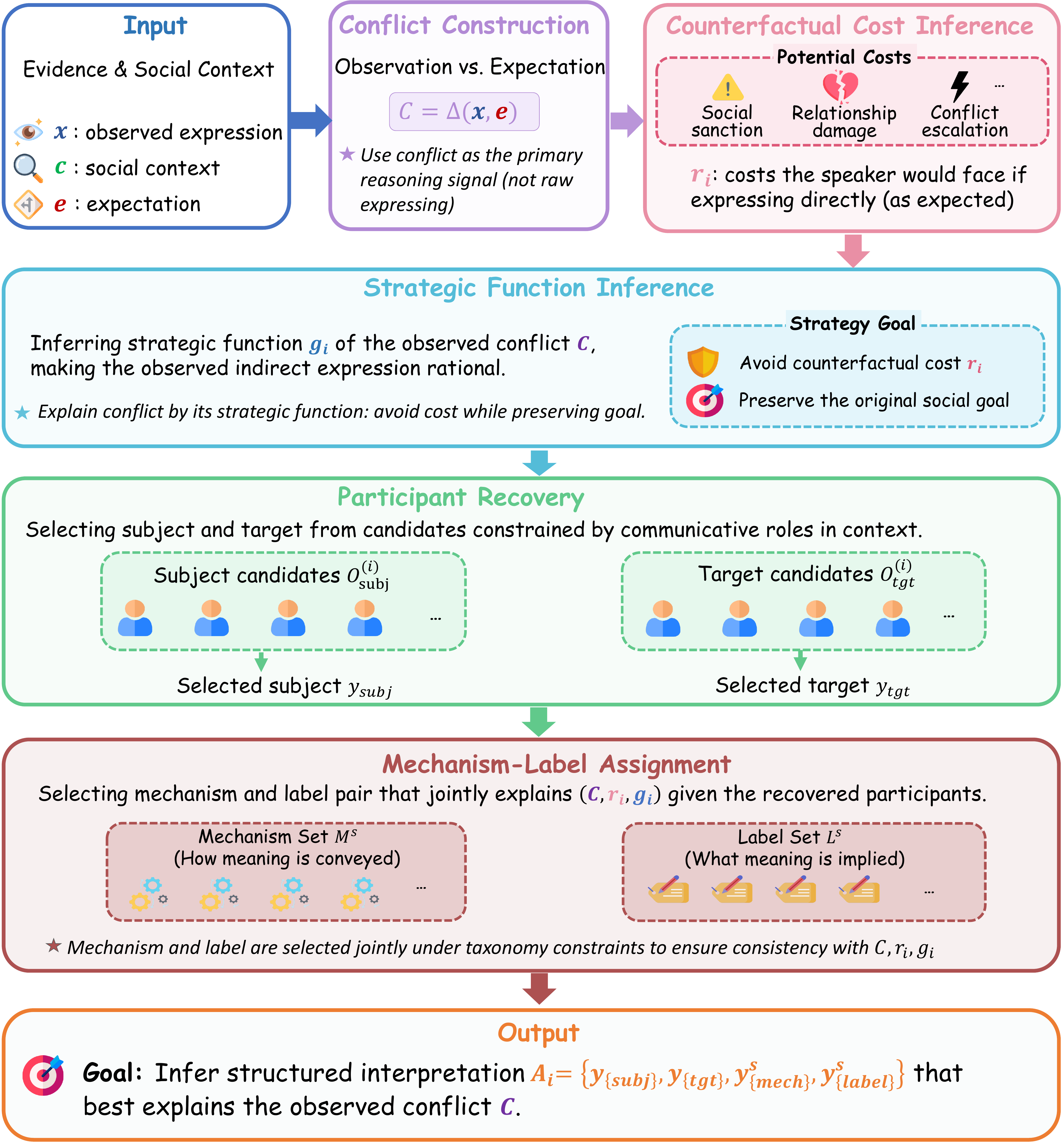}
    \caption{Illustration of the Conflict-Constrained Staged Abductive Inference mechanism.
    Building on the explicit perception $x$, social context $c$, and context-conditioned expectation $e$ established in the preceding stages, CoDAR constructs the conflict $C=\Delta(x,e)$ as the primary reasoning signal. CoDAR first infers the counterfactual cost $r_i$ of direct, expectation-aligned expression, and then derives the strategic function $g_i$ through which the observed conflict avoids this cost while preserving the subject's social goal. 
    These inferences guide participant recovery and scenario-specific mechanism--label assignment, yielding $A_i=\{y_\mathrm{{subj}},y_\mathrm{{tgt}},y^s_\mathrm{{mech}},y^s_\mathrm{{label}}\} \in O^{(i)}_\mathrm{{subj}} \times O_{\mathrm{tgt}}^{(i)} \times M_s \times L_s$}
    \label{fig:abductive}

  \end{figure}

  Because conflict $\mathcal{C}$ may admit several explanations, Step~4 restricts inference to MoCA's output schema and scenario-specific taxonomies.
  Figure~\ref{fig:abductive} summarizes the admissible tuple construction, staged abductive inference, and verification-guided revision.
  For instance $i$ in scenario $s$, let $\mathcal{O}^{(i)}_{\text{subj}}$ and $\mathcal{O}^{(i)}_{\text{tgt}}$ denote the provided subject and target candidate sets.
  Let $\mathcal{M}_{s}$ and $\mathcal{L}_{s}$ denote the scenario-specific mechanism and label sets.
  The admissible output space is
  \begin{equation}
      \Omega_{i,s}
      =
      \mathcal{O}^{(i)}_{\text{subj}}
      \times
      \mathcal{O}^{(i)}_{\text{tgt}}
      \times
      \mathcal{M}_{s}
      \times
      \mathcal{L}_{s}.
  \end{equation}
  CoDAR returns one structured tuple:
  \begin{equation}
      \mathcal{A}_{i}
      =
      \{y_{\text{subj}},y_{\text{tgt}},
      y^s_{\text{mech}},y^s_{\text{label}}\}
      \in \Omega_{i,s}.
  \end{equation}
  This constraint defines the form of $\mathcal{A}_{i}$; it does not enumerate or rank $\Omega_{i,s}$, learn a scorer, or introduce a separate decoding objective.
  
  \paratitle{Counterfactual Cost.}
  Given expectation $e$, CoDAR considers a direct, expectation-aligned expression and infers its likely social, relational, strategic, or power-related cost.
  This cost explains why the subject may prefer an implicit expression; no mechanism or label is assigned at this stage.
  
  \paratitle{Strategic Function of Conflict.}
  CoDAR next asks how the observed deviation $\mathcal{C}$ avoids this cost while preserving the subject's social goal.
  The inferred strategy must explain the observed multimodal deviation; $\mathcal{C}$ itself is not treated as the final label.
  
  \paratitle{Participant Recovery.}
  Using $x$, $c$, and $\mathcal{C}$, CoDAR selects one subject from $\mathcal{O}^{(i)}_{\text{subj}}$ and one target from $\mathcal{O}^{(i)}_{\text{tgt}}$.
  Selection follows communicative roles, not visual salience.
  
  \paratitle{Mechanism--Label Assignment.}
  CoDAR finally selects one mechanism from $\mathcal{M}_{s}$ and one label from $\mathcal{L}_{s}$.
  The mechanism specifies how the meaning is constructed or concealed, whereas the label identifies the latent social state.
  Their assignment must agree with the inferred cost, strategic function, participant relation, and observed conflict.
  
  The complete staged process can be summarized as
  \begin{equation}
  \begin{aligned}
      (x,c,e,\mathcal{C})
      &\longrightarrow
      \text{Counterfactual Cost} \\
      &\longrightarrow
      \text{Strategic Function of Conflict} \\
      &\longrightarrow
      (y_{\text{subj}},y_{\text{tgt}}) \\
      &\longrightarrow
      (y^s_{\text{mech}},y^s_{\text{label}}).
  \end{aligned}
  \end{equation}
  The final mapping uses the scenario taxonomy within prompt-guided inference; it is not a separately optimized mechanism-to-label decoder.
  The mechanism remains an explicit intermediate field linking $\mathcal{C}$ to the predicted label.
  
  \paratitle{Consistency Verification.}
  The verification function $\Phi$ checks three conditions:
  \begin{compactitem}
      \item \textit{\textbf{Explicit evidence consistency:}} $\mathcal{A}_{i}$ is supported by observable cues in $x$;
      \item \textit{\textbf{Social context consistency:}} $\mathcal{A}_{i}$ is compatible with the facts, connections, and norms in $c$;
      \item \textit{\textbf{Mechanism consistency:}} the mechanism explains $\mathcal{C}$ and its inferred strategic function.
  \end{compactitem}
  If all checks pass, $\mathcal{A}_{i}$ is accepted.
  Otherwise, the reported breakpoint triggers revision over Steps~2--5.
  
  Table~\ref{tab:ablation_results_merged} evaluates Steps~3 and 4 jointly.
  Removing them yields the largest overall degradation: \textit{Holistic Alignment} falls by 5.64 and 3.82 percentage points in Affection and Stance, respectively, while \textit{Mechanism Faithfulness} falls by 16.55 points in Intent.
  Because both steps are removed together, this ablation supports the combined unit but cannot attribute the gains to either step or to an individual inference stage.
  \section{Experiments}

  \subsection{Settings}
  
  \paratitle{Models.}
  We evaluate three groups of MLLMs under the same MoCA task: open-weight, affect-specialized, and proprietary models. Together, they support comparisons across model scale, reasoning variants, and domain specialization.
  
  \textbf{Open-weight MLLMs.}
  The open-weight baselines are LLaVA-OneVision-7B and LLaVA-OneVision-70B~\citep{li2024llavaonevisioneasyvisualtask}, InternVL3.5-38B~\citep{wang2025internvl35advancingopensourcemultimodal}, MiniCPM-V-4.5~\citep{yu2025minicpmv45cookingefficient}, and GLM-4.6V-Flash-9B~\citep{hong2025glm}. The two LLaVA-OneVision models enable a within-family comparison across scales. We additionally evaluate Qwen3-VL-32B-Instruct and Qwen3-VL-32B-Thinking~\citep{bai2025qwen3vltechnicalreport}. Their shared family and nominal scale provide a direct comparison between instruction-oriented and thinking-oriented variants, although the observed differences cannot be attributed solely to inference style.
  
  \textbf{Affect-specialized MLLMs.}
  To test whether explicit affect modeling transfers to implicit social reasoning, we evaluate AffectGPT~\citep{lian2025affectgpt} and Emotion-Qwen~\citep{huang2025emotionqwenunifiedframeworkemotion} under the same structured prediction objective.
  
  \textbf{Proprietary MLLMs.}
  We use Gemini-3.1-Pro~\citep{gemini2025}, Claude-Sonnet-4.6~\citep{claudeSonnet46SystemCard2026}, ChatGPT-5.4~\citep{gpt5systemcard2025}, and ChatGPT-4o~\citep{gpt4o2024} as proprietary capability references. Their undisclosed architectures and training data preclude controlled architectural attribution.
  
  \paratitle{Input Processing.}
  Each image-text instance is presented as the original caption and its paired image, preserving the evidence required to resolve cross-modal agreement or conflict. Each video is uniformly sampled into a temporally ordered frame sequence, while the original textual input remains available to the model. The reported video results therefore measure reasoning over sampled temporal evidence rather than uninterrupted raw-video perception. Speech is transcribed into captions that retain linguistic content and speaking style. These transcripts are combined with the corresponding visual evidence. Audio results therefore measure transcript-conditioned rather than direct acoustic reasoning.
  
  \paratitle{Evaluation Protocol.}
  Every model predicts the same four fields: Subject, Target, Mechanism, and Label.
  The normalization, invalid-output rule, and four metrics defined in Section~\ref{sec:evaluation} are applied unchanged to model and human responses.

  \subsection{Main Results and Observations}
  \label{sec:main_results}
  
  Tables~\ref{tab:results_affection_full}--\ref{tab:results_stance_full} report all scenario results.
  \begin{table}[!t]
  \centering
  \caption{Performance (accuracy) in the Affection Scenario. 
  Lab (Label), Subj (Subject), Tgt (Target), Mech (Mechanism), Dyad (Dyadic Alignment), Faith (Mechanism Faithfulness), and Holis (Holistic Alignment). The best results within each category are in \colorbox{mygreen}{\textcolor{black}{green}}.
  \colorbox{myblue}{\textcolor{black}{Light blue}} highlights human performance, while \colorbox{imppurple}{\textcolor{black}{light purple}} indicates CoDAR-augmented models.
  \textcolor{imppos}{+} and \textcolor{impneg}{-} indicate CoDAR's performance improvement and degradation over the backbones, respectively.
  }
  \label{tab:results_affection_full}
  \fontsize{8.5pt}{9.8pt}\selectfont
  \setlength{\tabcolsep}{4pt}
  \begin{tabular*}{\textwidth}{@{\extracolsep{\fill}}lccccccc@{}}
  \toprule
  \textbf{Method} & \textbf{Lab} & \textbf{Subj} & \textbf{Tgt} & \textbf{Mech} & \textbf{Dyad} & \textbf{Faith} & \textbf{Holis} \\
  \midrule
  \multicolumn{8}{c}{$\bullet$ \textit{\textbf{Human}}} \\
  \rowcolor{myblue} Human & 80.09 & 88.50 & 90.04 & 72.12 & 78.54 & 77.90 & 59.73 \\
  $\Delta$ (Human-Best) & 31.11 & 2.39 & 4.72 & 22.71 & 3.19 & 8.74 & 41.89 \\
  
  \hline
  \multicolumn{8}{c}{$\bullet$ \textit{\textbf{Open-weight Models}}} \\
  LLaVA-{\scriptsize OneVision-7B} & 23.29 & 38.14 & 45.41 & 23.50 & 20.94 & 27.98 & 1.50 \\
  
  GLM-{\scriptsize 4.6V-Flash-9B} & 36.00 & 70.83 & 72.12 & 39.96 & 53.95 & 41.84 & 8.12 \\
  MiniCPM-{\scriptsize V-4.5} & 36.32 & 46.05 & 65.81 & 36.75 & 32.91 & 38.53 & 4.91 \\
  Qwen3-{\scriptsize VL-32B-Thinking} & 35.90 & 67.09 & 77.56 & 24.89 & 53.95 & 25.30 & 5.24 \\
  LLaVA-{\scriptsize OneVision-70B} & 34.73 & 73.39 & 77.92 & 31.38 & 58.47 & 36.43 & 6.32 \\
  InternVL3.5-{\scriptsize 38B} & 36.97 & 70.19 & 72.97 & 41.67 & 52.99 & 46.53  & 10.04 \\
  \rowcolor{imppurple} \textbf{CoDAR}+{\scriptsize InternVL3.5-38B} & 38.57 & 80.98 & \cellcolor{mygreen}81.41 & 42.31 & 62.39 & 55.12 & \cellcolor{mygreen}13.78 \\
  \specialrule{0em}{-2pt}{-1pt} &  \inc{1.60} & \inc{10.79} & \inc{8.44} & \inc{0.64} & \inc{9.40} & \inc{8.59} & \inc{3.74} \\
  
  Qwen3-{\scriptsize VL-32B-Instruct} & 39.53 & 78.42 & 77.03 & 40.81 & 61.75 & 41.62 & 9.08 \\
  \rowcolor{imppurple} \textbf{CoDAR}+{\scriptsize Qwen3-VL-32B-Inst} & \cellcolor{mygreen}40.28 & \cellcolor{mygreen}83.76 & 79.49 & \cellcolor{mygreen}47.22 & \cellcolor{mygreen}64.74 & \cellcolor{mygreen}59.68 & 11.54 \\
  \specialrule{0em}{-2pt}{-1pt} & \inc{0.75}  & \inc{5.34}  & \inc{2.46}  &  \inc{6.41} &  \inc{2.99} &  \inc{18.06} &  \inc{2.46} \\
  
  \hline
  \multicolumn{8}{c}{$\bullet$ \textit{\textbf{Open-weight Affective-specific Models}}} \\
  AffectGPT & 26.82 & 6.84 & 21.47 & 43.48 & 3.95 & 37.85 & 0.64 \\
  \rowcolor{imppurple} \textbf{CoDAR}+{\scriptsize AffectGPT} & \cellcolor{mygreen}31.84 & 11.54 & 23.40 & \cellcolor{mygreen}44.23 & 6.73 & \cellcolor{mygreen}38.59 & \cellcolor{mygreen}2.88 \\
  \specialrule{0em}{-2pt}{-1pt} & \inc{5.02} & \inc{4.70} & \inc{1.93} & \inc{0.75} & \inc{2.78} & \inc{0.74} & \inc{2.24} \\
  Emotion-{\scriptsize Qwen} & 27.88 & 26.28 & 34.40 & 27.56 & 15.71 & 29.50 & 0.75 \\
  \rowcolor{imppurple} \textbf{CoDAR}+{\scriptsize Emotion-Qwen} & 31.09 & \cellcolor{mygreen}28.74 & \cellcolor{mygreen}37.07 & 29.17 & \cellcolor{mygreen}17.63 & 31.27 & 1.82 \\
  \specialrule{0em}{-2pt}{-1pt} & \inc{3.21} & \inc{2.46} & \inc{2.67} & \inc{1.61} & \inc{1.92} & \inc{1.77} & \inc{1.07} \\
  
  \hline
  \multicolumn{8}{c}{$\bullet$ \textit{\textbf{Proprietary Models}}} \\
  ChatGPT-{\scriptsize 4o} & 41.60 & 77.54 & 78.18 & \cellcolor{mygreen}49.41 & 61.39 & 48.33 & 12.51 \\
  Claude-{\scriptsize Sonnet-4.6} & 43.47 & 76.05 & 78.32 & 43.80 & 61.60 & 44.17 & 12.19 \\
  
  ChatGPT-{\scriptsize 5.4} & 44.02 & 81.84 & 81.62 & 41.77 & 70.51 & 42.48 & 12.71 \\
  \rowcolor{imppurple} \textbf{CoDAR}+{\scriptsize ChatGPT-5.4} & 45.73 & \cellcolor{mygreen}86.11 & 82.26 & 43.59 & 72.01 & \cellcolor{mygreen}69.16 & \cellcolor{mygreen}17.84 \\
  \specialrule{0em}{-2pt}{-1pt} & \inc{1.71} & \inc{4.27}   & \inc{0.64}   &  \inc{1.82}  & \inc{1.50}   & \inc{26.68}   &  \inc{5.13} \\
  
  Gemini-{\scriptsize 3.1-Pro} & 45.34 & 80.28 & 82.96 & 42.12 & 74.92 & 43.97 & 14.58 \\
  \rowcolor{imppurple} \textbf{CoDAR}+{\scriptsize Gemini-3.1-Pro} & \cellcolor{mygreen}48.98 & 81.78 & \cellcolor{mygreen}85.32 & 47.27 & \cellcolor{mygreen}75.35 & 66.52 & 17.47 \\
  \specialrule{0em}{-2pt}{-1pt} & \inc{3.64} &  \inc{1.50}  &  \inc{2.36} &\inc{5.15}&  \inc{0.43}  & \inc{22.55}   & \inc{2.89} \\
  
  \bottomrule
  \end{tabular*}
  \end{table}

  \paratitle{Performance Disparity Across Dimensions}. A consistent pattern is observed across all scenarios: model performance remains relatively strong on dyadic perception but decreases substantially on reasoning-related dimensions. Comparing the best-performing models including CoDAR to human performance, the gap on Dyadic Alignment remains limited (3.19\%--7.46\%), indicating that current MLLMs can reliably identify interaction participants. The gap increases on Label accuracy (31.04\%--34.35\%), which requires recovering implicit social states, and further enlarges on Holistic Alignment (37.38\%--41.89\%), where all components must be correct simultaneously. This decline indicates that MoCA's main difficulty is recovering the communicative configuration, not identifying participants.

  \paratitle{Model Performance Comparison}. Proprietary models consistently outperform open-weight models, although the magnitude varies across evaluation dimensions.
  On Subject and Target, strong open-weight models such as Qwen3-VL-32B-Instruct achieve performance comparable to proprietary systems.
  But on Holistic Alignment, the gap widens: the best proprietary baseline (24.69\%) exceeds the best open-weight baseline (15.71\%) by nearly 10 percentage points in Stance, yet both remain far below human performance (67.64\%).
  Affect-specialized models do not compensate for this gap. AffectGPT achieves 43.48\% Mechanism accuracy in Affection but only 3.95\% Dyadic Alignment and 0.64\% Holistic Alignment, indicating that explicit emotion recognition does not translate to improved implicit relational understanding.
  Qwen3-VL-32B-Thinking also performs below its Instruct variant in Affection (5.24\% versus 9.08\%), suggesting that explicit reasoning traces do not necessarily improve consistency between predicted labels and mechanisms.

  \paratitle{Effect of CoDAR}.
  CoDAR yields consistent improvements across model families, with the largest gains in Mechanism Faithfulness and Holistic Alignment. We apply CoDAR to representative models from each category; Intent additionally includes LLaVA-OneVision-70B because of its strong baseline, yielding three evaluated models. In Affection, Mechanism Faithfulness increases from 42.48\% to 69.16\% for ChatGPT-5.4 and from 41.62\% to 59.68\% for Qwen3-VL-32B-Instruct.

  Table~\ref{tab:results_affection_full} further localizes the Affection bottleneck. Relative to human performance, the best machine results are only 2.39 and 4.72 percentage points lower on Subject and Target, respectively, but the gaps widen to 31.11 points on Label and 41.89 points on Holistic Alignment.

  The same separation appears within individual baselines. Gemini-3.1-Pro reaches 80.28\% Subject and 82.96\% Target accuracy, but only 14.58\% Holistic Alignment. AffectGPT attains 43.48\% Mechanism accuracy while obtaining 3.95\% Dyadic Alignment and 0.64\% Holistic Alignment. These values show that recovering one participant or mechanism does not by itself satisfy the joint four-output requirement. CoDAR's changes reinforce this distinction. For Qwen3-VL-32B-Instruct, it improves Mechanism Faithfulness by 18.06 points while raising Target and Holistic Alignment by 2.46 points each; for ChatGPT-5.4, the corresponding gains are 26.68, 0.64, and 5.13 points. Thus, the principal improvement lies in making predicted labels more consistent with their mechanisms, rather than uniformly increasing every component. Nevertheless, the best machine Holistic Alignment remains 17.84\%, compared with 59.73\% for human annotators.

  \begin{table}[!t]
  \centering
  \caption{Intent-scenario accuracy; notation and metric abbreviations follow Table~\ref{tab:results_affection_full}.}
  \label{tab:results_intent_full}
  \fontsize{8.5pt}{9.8pt}\selectfont
  \setlength{\tabcolsep}{4pt}
  {\renewcommand{\arraystretch}{1.04}
  \begin{tabular*}{\textwidth}{@{\extracolsep{\fill}}lccccccc@{}}
  \toprule
  \textbf{Method} & \textbf{Lab} & \textbf{Subj} & \textbf{Tgt} & \textbf{Mech} & \textbf{Dyad} & \textbf{Faith} & \textbf{Holis} \\
  \midrule
  \multicolumn{8}{c}{$\bullet$ \textit{\textbf{Human}}} \\
  \rowcolor{myblue} Human & 77.27 & 92.67 & 91.27 & 82.4 & 85.33 & 86.19 & 65.20 \\
  $\Delta$ (Human-Best) & 31.04 & 2.93 & 4.91 & 9.20 & 7.46 & 1.65 & 37.38 \\
  
  \hline
  \multicolumn{8}{c}{$\bullet$ \textit{\textbf{Open-weight Models}}} \\
  LLaVA-{\scriptsize OneVision-7B} & 16.62 & 43.43 & 47.58 & 13.57 & 22.90 & 44.90 & 1.27 \\
  
  GLM-{\scriptsize 4.6V-Flash-9B} & 29.85 & 58.69 & 71.84 & 70.23 & 42.15 & \cellcolor{mygreen}82.38 & 10.34 \\
  MiniCPM-{\scriptsize V-4.5} & 33.25 & 51.06 & 69.30 & 31.21 & 36.30 & 38.52 & 5.94 \\
  Qwen3-{\scriptsize VL-32B-Thinking} & 20.70 & 63.70 & 78.88 & 22.22 & 51.31 & 43.44 & 5.43 \\
  
  LLaVA-{\scriptsize OneVision-70B} & 29.75 & 75.98 & 71.75 & 69.97 & 58.14 & 81.04 & 14.38 \\
  \rowcolor{imppurple} \textbf{CoDAR}+{\scriptsize LLaVA-OneVision-70B} & 34.21 & \cellcolor{mygreen}84.80 & 77.98 & \cellcolor{mygreen}71.97 & 66.88 & 81.91 & \cellcolor{mygreen}21.29 \\
  \specialrule{0em}{-2pt}{-1pt} & \inc{4.46} & \inc{8.82} & \inc{6.23} & \inc{2.00} & \inc{8.74} & \inc{0.87} & \inc{6.91} \\
  
  InternVL3.5-{\scriptsize 38B} & 31.21 & 67.09 & 71.67 & 65.06 & 49.28 & 77.72 & 11.37 \\
  \rowcolor{imppurple} \textbf{CoDAR}+{\scriptsize InternVL3.5-38B} & 34.94 & 84.14 & \cellcolor{mygreen}83.63 & 65.90 & 69.30 & 79.13 & 19.08 \\
  \specialrule{0em}{-2pt}{-1pt} & \inc{3.73} & \inc{17.05} & \inc{11.96} & \inc{0.84} & \inc{20.02} & \inc{1.41} & \inc{7.71} \\
  
  Qwen3-{\scriptsize VL-32B-Instruct} & 27.82 & 73.28 & 75.06 & 50.98 & 54.88 & 65.85 & 10.6 \\
  \rowcolor{imppurple} \textbf{CoDAR}+{\scriptsize Qwen3-VL-32B-Instruct} & \cellcolor{mygreen}38.00 & 84.22 & 80.92 & 67.68 & \cellcolor{mygreen}69.47 & 75.89 & 19.93 \\
  \specialrule{0em}{-2pt}{-1pt} & \inc{10.18}  & \inc{10.94}  & \inc{5.86}  &  \inc{16.70} &  \inc{14.59} &  \inc{10.04} &  \inc{9.33} \\
  
  \hline
  \multicolumn{8}{c}{$\bullet$ \textit{\textbf{Proprietary Models}}} \\
  ChatGPT-{\scriptsize 4o} & 36.65 & 67.81 & 78.88 & 72.53 & 53.56 & \cellcolor{mygreen}84.54 & 15.36 \\
  Claude-{\scriptsize Sonnet-4.6} & 37.85 & 75.11 & 80.26 & 70.21 & 61.63 & 80.05 & 17.94 \\
  
  ChatGPT-{\scriptsize 5.4} & 38.51 & 80.15 & 82.36 & 71.08 & 66.75 & 81.50 & 20.02 \\
  \rowcolor{imppurple} \textbf{CoDAR}+{\scriptsize ChatGPT-5.4} & \cellcolor{mygreen}46.23 & \cellcolor{mygreen}89.74 & 83.04 & \cellcolor{mygreen}73.20 & 75.40 & 82.02 & \cellcolor{mygreen}27.82 \\
  \specialrule{0em}{-2pt}{-1pt} & \inc{7.72} & \inc{9.59}   & \inc{0.68}   &  \inc{2.12}  & \inc{8.65}   & \inc{0.52}   &  \inc{7.80} \\
  
  Gemini-{\scriptsize 3.1-Pro} & 35.42 & 86.79 & 84.65 & 71.44 & 75.47 & 77.48 & 19.90 \\
  \rowcolor{imppurple} \textbf{CoDAR}+{\scriptsize Gemini-3.1-Pro} & 46.05 & 88.85 & \cellcolor{mygreen}86.36 & 72.04 & \cellcolor{mygreen}77.87 & 78.03 & 26.33 \\
  \specialrule{0em}{-2pt}{-1pt} & \inc{10.63} &  \inc{2.06}  &  \inc{1.71} & \inc{0.60} &  \inc{2.40}  & \inc{0.55}   & \inc{6.43} \\
  
  \bottomrule
  \end{tabular*}
  }
  \end{table}

  In Intent, Holistic Alignment improves by 7.80\% for ChatGPT-5.4 and 9.33\% for Qwen3-VL-32B-Instruct. Table~\ref{tab:results_intent_full} also shows why component-level scores alone are insufficient: GLM-4.6V-Flash-9B obtains 70.23\% Mechanism accuracy and 82.38\% Mechanism Faithfulness, yet only 10.34\% Holistic Alignment; ChatGPT-4o reaches 72.53\%, 84.54\%, and 15.36\% on the same metrics. Across all five reported CoDAR pairs, Holistic Alignment increases by 6.43--9.33 points, despite substantial variation in the accompanying atomic-component gains. The consistent improvement in the joint metric therefore reflects better coordination of the four required outputs, rather than strength on one isolated component.

  These gains indicate that organizing reasoning around conflict identification and abductive explanation helps bridge the gap between surface perception and mechanism-level understanding. Subject and Target performance remains stable, confirming that improvement originates from deeper reasoning rather than perceptual enhancement.

  \begin{table}[!t]
  \centering
  \caption{Stance-scenario accuracy; notation and metric abbreviations follow Table~\ref{tab:results_affection_full}.
  }
  \label{tab:results_stance_full}
  \fontsize{8.5pt}{9.8pt}\selectfont
  \setlength{\tabcolsep}{4pt}
  {\renewcommand{\arraystretch}{1.04}
  \begin{tabular*}{\textwidth}{@{\extracolsep{\fill}}lccccccc@{}}
  \toprule
  
  \textbf{Method} & \textbf{Lab} & \textbf{Subj} & \textbf{Tgt} & \textbf{Mech} & \textbf{Dyad} & \textbf{Faith} & \textbf{Holis} \\
  \midrule
  \multicolumn{8}{c}{$\bullet$ \textit{\textbf{Human}}} \\
  \rowcolor{myblue} Human & 79.96 & 91.86 & 92.90 & 79.54 & 84.76 & 84.86 & 67.64 \\
  $\Delta$ (Human-Best) & 34.35 & 1.55 & 5.99 & 8.55 & 6.39 & 2.37 & 41.05 \\
  
  \hline
  
  \multicolumn{8}{c}{$\bullet$ \textit{\textbf{Open-weight Models}}} \\
  LLaVA-{\scriptsize OneVision-7B} & 20.44 & 39.17 & 50.25 & 47.03 & 20.64 & 56.65 & 2.01 \\
  
  GLM-{\scriptsize 4.6V-Flash-9B} & 30.01 & 69.99 & 70.49 & 59.21 & 50.25 & 70.81 & 11.18 \\
  MiniCPM-{\scriptsize V-4.5} & 17.52 & 52.37 & 69.49 & 31.62 & 37.87 & 33.91 & 2.42 \\
  Qwen3-{\scriptsize VL-32B-Thinking} & 35.05 & 66.47 & 78.55 & 62.84 & 52.27 & 70.69 & 12.08 \\
  LLaVA-{\scriptsize OneVision-70B} & 23.57 & 75.55 & 77.96 & 58.88 & 59.76 & 65.58 & 9.54 \\
  InternVL3.5-{\scriptsize 38B} & 32.53 & 72.31 & 75.43 & 60.32 & 54.58 & 69.66 & 11.68 \\
  \rowcolor{imppurple} \textbf{CoDAR}+{\scriptsize InternVL3.5-38B} & 33.85 & 84.69 & \cellcolor{mygreen}86.91 & 60.62 & \cellcolor{mygreen}71.60 & \cellcolor{mygreen}81.55 & 18.33 \\
  \specialrule{0em}{-2pt}{-1pt} & \inc{1.32} & \inc{12.38} & \inc{11.48} & \inc{0.30} & \inc{17.02} & \inc{11.89} & \inc{6.65} \\
  Qwen3-{\scriptsize VL-32B-Instruct} & 33.84 & 77.64 & 78.45 & 68.88 & 60.62 & 77.38 & 15.71 \\
  \rowcolor{imppurple} \textbf{CoDAR}+{\scriptsize Qwen3-VL-32B-Instruct} & \cellcolor{mygreen}37.26 & \cellcolor{mygreen}86.00 & 79.86 & \cellcolor{mygreen}69.28 & 66.77 & 78.38 & \cellcolor{mygreen}20.64 \\
  \specialrule{0em}{-2pt}{-1pt} & \inc{3.42}  & \inc{8.36}  & \inc{1.41}  &  \inc{0.40} &  \inc{6.15} &  \inc{1.00} &  \inc{4.93} \\
  \hline
  
  \multicolumn{8}{c}{$\bullet$ \textit{\textbf{Proprietary Models}}} \\
  ChatGPT-{\scriptsize 4o} & 38.23 & 75.13 & 79.20 & 70.44 & 59.73 & 77.07 & 17.13 \\
  Claude-{\scriptsize Sonnet-4.6} & 37.12 & 80.12 & 78.90 & \cellcolor{mygreen}70.99 & 65.52 & 79.51 & 20.59 \\
  ChatGPT-{\scriptsize 5.4} & 38.57 & 86.30 & 80.77 & 67.27 & 70.39 & 81.46 & 21.95 \\
  \rowcolor{imppurple} \textbf{CoDAR}+{\scriptsize ChatGPT-5.4} & 41.99 & 88.62 & 84.19 & 68.68 & 74.62 & \cellcolor{mygreen}82.49 & \cellcolor{mygreen}26.59 \\
  \specialrule{0em}{-2pt}{-1pt} & \inc{3.42} & \inc{2.32}   & \inc{3.42}   &  \inc{1.41}  & \inc{4.23}   & \inc{1.03}   &  \inc{4.64} \\
  Gemini-{\scriptsize 3.1-Pro} & 42.65 & 89.59 & 84.18 & 69.39 & 76.84 & 75.60 & 24.69 \\
  \rowcolor{imppurple} \textbf{CoDAR}+{\scriptsize Gemini-3.1-Pro} & \cellcolor{mygreen}45.61 & \cellcolor{mygreen}90.31 & \cellcolor{mygreen}85.71 & 70.20 & \cellcolor{mygreen}78.37 & 76.73 & 25.92 \\
  \specialrule{0em}{-2pt}{-1pt} & \inc{2.96} &  \inc{0.72}  &  \inc{1.53} &\inc{0.81}&  \inc{1.53}  & \inc{1.13}   & \inc{1.23} \\
  \bottomrule
  \end{tabular*}
  }
  \end{table}

  Table~\ref{tab:results_stance_full} exhibits the same separation between component recovery and complete structured prediction. Gemini-3.1-Pro is the strongest baseline on Holistic Alignment at 24.69\%, but remains 42.95 points below human performance. CoDAR raises Holistic Alignment by 6.65 points for InternVL3.5-38B, 4.93 points for Qwen3-VL-32B-Instruct, and 4.64 points for ChatGPT-5.4. For InternVL3.5-38B, this improvement accompanies a 17.02-point increase in Dyadic Alignment but only a 0.30-point change in Mechanism accuracy. The contrast further shows that recovering a coherent Subject--Target relation and integrating it with the predicted mechanism and label are distinct requirements of the task.

  \subsection{Ablation Analysis}
  To verify the contribution of each step in CoDAR, we design three ablation settings:
  \textbf{1) w/o Step 2:} removing the social context construction step;
  \textbf{2) w/o Step 3\&4:} removing both the expectation-and-conflict modeling step and the abductive reasoning step, which together constitute the core reasoning unit of CoDAR;
  \textbf{3) w/o Step 5:} removing the consistency verification step.
  We evaluate all variants across the three scenarios, with the results reported in Table~\ref{tab:ablation_results_merged}.

  \begin{table}[!t]
  \centering
  \caption{Ablation study results across three scenarios based on Qwen3-VL-32B-Instruct. We report the average of the four atomic metrics (\textit{Lab}, \textit{Subj}, \textit{Tgt}, \textit{Mech}) as \textbf{Atomic Avg.}, together with \textbf{Dyad}, \textbf{Faith}, and \textbf{Holis}. Green indicates drops relative to full CoDAR, while red indicates gains.}
  \label{tab:ablation_results_merged}
  \fontsize{8.5pt}{9.8pt}\selectfont
  \setlength{\tabcolsep}{4pt}
  {\renewcommand{\arraystretch}{1.0}
  \begin{tabular*}{\textwidth}{@{\extracolsep{\fill}}lrrrr@{}}
  \toprule
  \textbf{Method} & \textbf{Atomic Avg.} & \textbf{Dyad} & \textbf{Faith} & \textbf{Holis} \\
  \midrule
  
  \multicolumn{5}{c}{\textbf{• Affection}} \\
  \textbf{CoDAR} 
  & 62.69
  & 64.74
  & 59.68
  & 11.54 \\
  \quad w/o Step 2   
  & 52.37 \dec{10.32}
  & 43.06 \dec{21.68}
  & 59.10 \dec{0.58}
  & 8.84 \dec{2.70} \\
  \quad w/o Step 3\&4 
  & 49.08 \dec{13.61}
  & 36.76 \dec{27.98}
  & 55.35 \dec{4.33}
  & 5.90 \dec{5.64} \\
  \quad w/o Step 5   
  & 50.16 \dec{12.53}
  & 45.05 \dec{19.69}
  & 57.62 \dec{2.06}
  & 8.69 \dec{2.85} \\
  
  \midrule
  \multicolumn{5}{c}{\textbf{• Intent}} \\
  \textbf{CoDAR} 
  & 67.71
  & 69.47
  & 75.89
  & 19.93 \\
  \quad w/o Step 2   
  & 62.39 \dec{5.32}
  & 78.00 \inc{8.53}
  & 64.92 \dec{10.97}
  & 15.68 \dec{4.25} \\
  \quad w/o Step 3\&4 
  & 54.55 \dec{13.16}
  & 76.19 \inc{6.72}
  & 59.34 \dec{16.55}
  & 15.22 \dec{4.71} \\
  \quad w/o Step 5   
  & 60.24 \dec{7.47}
  & 75.66 \inc{6.19}
  & 63.53 \dec{12.36}
  & 15.61 \dec{4.32} \\
  
  \midrule
  \multicolumn{5}{c}{\textbf{• Stance}} \\
  \textbf{CoDAR} 
  & 68.10
  & 66.77
  & 78.38
  & 20.64 \\
  \quad w/o Step 2   
  & 62.34 \dec{5.76}
  & 75.00 \inc{8.23}
  & 66.06 \dec{12.32}
  & 18.73 \dec{1.91} \\
  \quad w/o Step 3\&4 
  & 58.03 \dec{10.07}
  & 71.06 \inc{4.29}
  & 63.34 \dec{15.04}
  & 16.82 \dec{3.82} \\
  \quad w/o Step 5   
  & 61.78 \dec{6.32}
  & 73.62 \inc{6.85}
  & 63.95 \dec{14.43}
  & 17.52 \dec{3.12} \\
  
  \bottomrule
  \end{tabular*}
  }
  \end{table}
  
  Overall, the full CoDAR model achieves the most stable performance, and all ablated variants perform worse on \textit{Holistic Alignment}. This confirms that CoDAR's gains arise from the complete reasoning loop rather than any single step in isolation.
  Among all ablations, w/o Step 3\&4 causes the most substantial and consistent degradation across all metrics, with \textit{Holistic Alignment} dropping by 5.64\% in Affection and 3.82\% in Stance, confirming that contextual expectation construction, conflict modeling, and abductive explanation form the central source of CoDAR's advantage. An interesting pattern emerges in Intent and Stance, where all ablated variants show higher \textit{Dyadic Alignment} than the full model. 
  However, this improvement co-occurs with sharp drops in \textit{Mechanism Faithfulness} (up to 16.55\% in Intent) and \textit{Holistic Alignment}, suggesting that without the full reasoning chain, models fall back on surface-level relational cues to match \textit{Subject} and \textit{Target}.

  \FloatBarrier

  \begin{figure}[!t]
  \centering
  \includegraphics[width=0.94\textwidth]{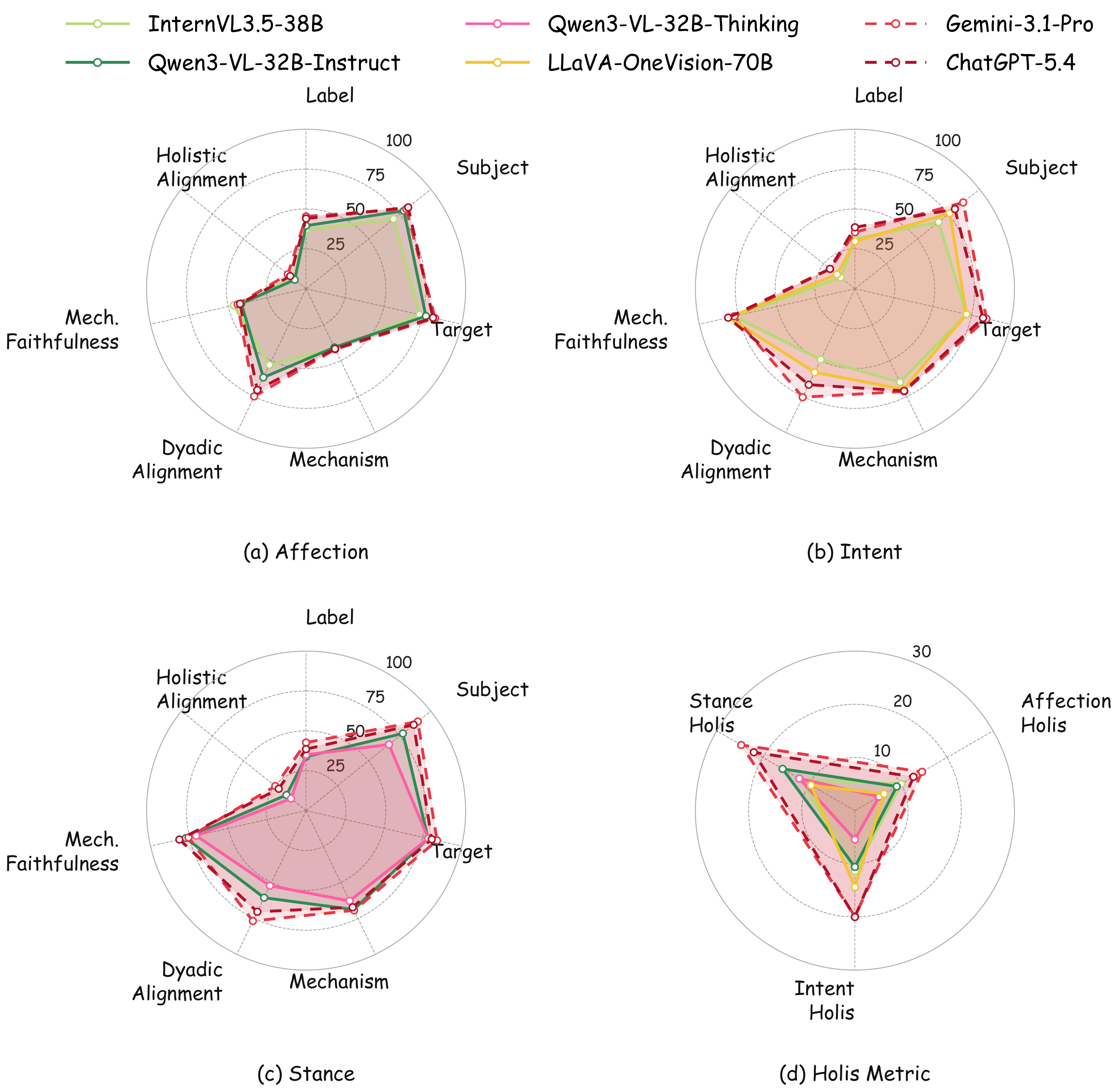}
  \caption{Baseline radar comparison across three scenarios with a holistic metric summary. Solid lines indicate open-weight models, while dashed lines indicate proprietary models. Colors are consistent for each model across subplots.
  }
  \label{fig:scenario_radar}
  \end{figure}
  
  \section{Analysis and Discussion}

  \subsection{Performance across Evaluation Metrics}
  
  \paratitle{\textit{RQ-1: Where Does the Performance Gap between Open-Weight and Proprietary MLLMs Emerge?}}\label{sec:rq1_analysis}
  Figure~\ref{fig:scenario_radar}a--c and Tables~\ref{tab:results_affection_full}--\ref{tab:results_stance_full} show that the gap is uneven across the seven metrics. Within the six plotted baselines, the strongest open-weight and proprietary \textit{Mechanism} scores remain close in all scenarios. Their gaps are only 0.45, 1.47, and 0.51 percentage points in Affection, Intent, and Stance, respectively. \textit{Mechanism Faithfulness} is similarly close in Intent, at 81.04\% versus 81.50\%, and Stance, at 77.38\% versus 81.46\%. In Affection, InternVL3.5-38B instead exceeds the best plotted proprietary baseline on this metric, at 46.53\% versus 43.97\%. The corresponding \textit{Label} gaps are 5.81, 7.30, and 7.60 points, while the \textit{Target} gaps are 5.04, 5.77, and 5.63 points. The \textit{Subject} gap increases from 3.42 points in Affection to 10.81 and 11.95 points in Intent and Stance.

  The separation is larger and consistent on the joint metrics. The strongest plotted proprietary baseline exceeds the strongest open-weight baseline on \textit{Dyadic Alignment} by 13.17, 17.33, and 16.22 points across the three scenarios. The corresponding \textit{Holistic Alignment} gaps are 4.54, 5.64, and 8.98 points, with the largest separation in Stance (Figure~\ref{fig:scenario_radar}d). However, even the strongest proprietary baseline remains 45.15, 45.18, and 42.95 points below human \textit{Holistic Alignment}, respectively. These comparisons locate the model-family gap primarily in joint relational and four-component consistency, rather than uniform superiority on every atomic component.
  \FloatBarrier
  \begin{figure}[!t]
      \centering
      \includegraphics[width=\textwidth]{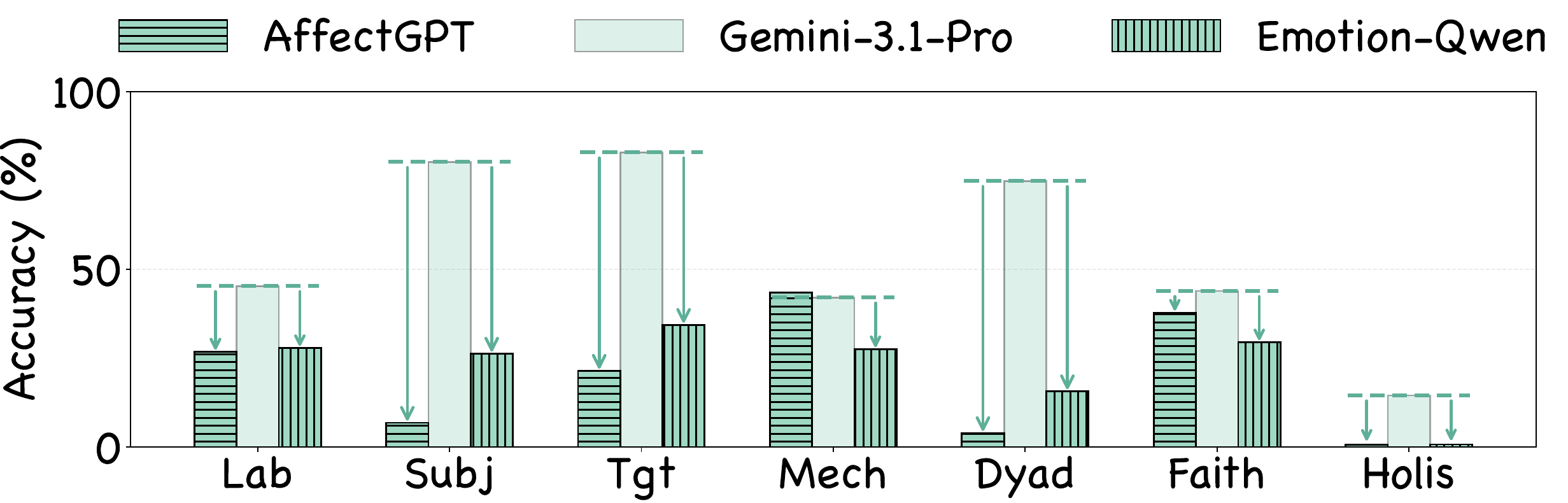}
      \caption{Comparison of AffectGPT, Emotion-Qwen, and Gemini-3.1-Pro on seven metrics in the Affection scenario. Horizontally hatched bars denote AffectGPT, vertically hatched bars denote Emotion-Qwen, and lightly shaded bars denote Gemini-3.1-Pro.
      }
      \label{fig:rq2_affective_vs_baseline}
  \end{figure}

  \paratitle{\textit{RQ-2: Can Affect-Specialized Models Excel in the Affection Scenario?}}\label{sec:rq2_analysis}
  Figure~\ref{fig:rq2_affective_vs_baseline} and Table~\ref{tab:results_affection_full} show that neither affect-specialized baseline matches Gemini-3.1-Pro on relational or holistic recovery. AffectGPT reaches 43.48\% \textit{Mechanism} accuracy, compared with 42.12\% for Gemini-3.1-Pro, but records only 3.95\% \textit{Dyadic Alignment} and 0.64\% \textit{Holistic Alignment}. Emotion-Qwen records 15.71\% \textit{Dyadic Alignment}, yet its \textit{Holistic Alignment} remains 0.75\%. By comparison, Gemini-3.1-Pro obtains 74.92\% and 14.58\% on these two metrics, respectively.

  Table~\ref{tab:results_affection_full} further shows that CoDAR raises \textit{Holistic Alignment} from 0.64\% to 2.88\% for AffectGPT and from 0.75\% to 1.82\% for Emotion-Qwen. These low absolute scores show that component-level strength does not ensure directed Subject--Target recovery or four-output alignment. This finding applies only to these two models and does not rule out broader benefits from affect specialization.

  \begin{figure}[!t]
    \centering
    \includegraphics[width=0.92\textwidth]{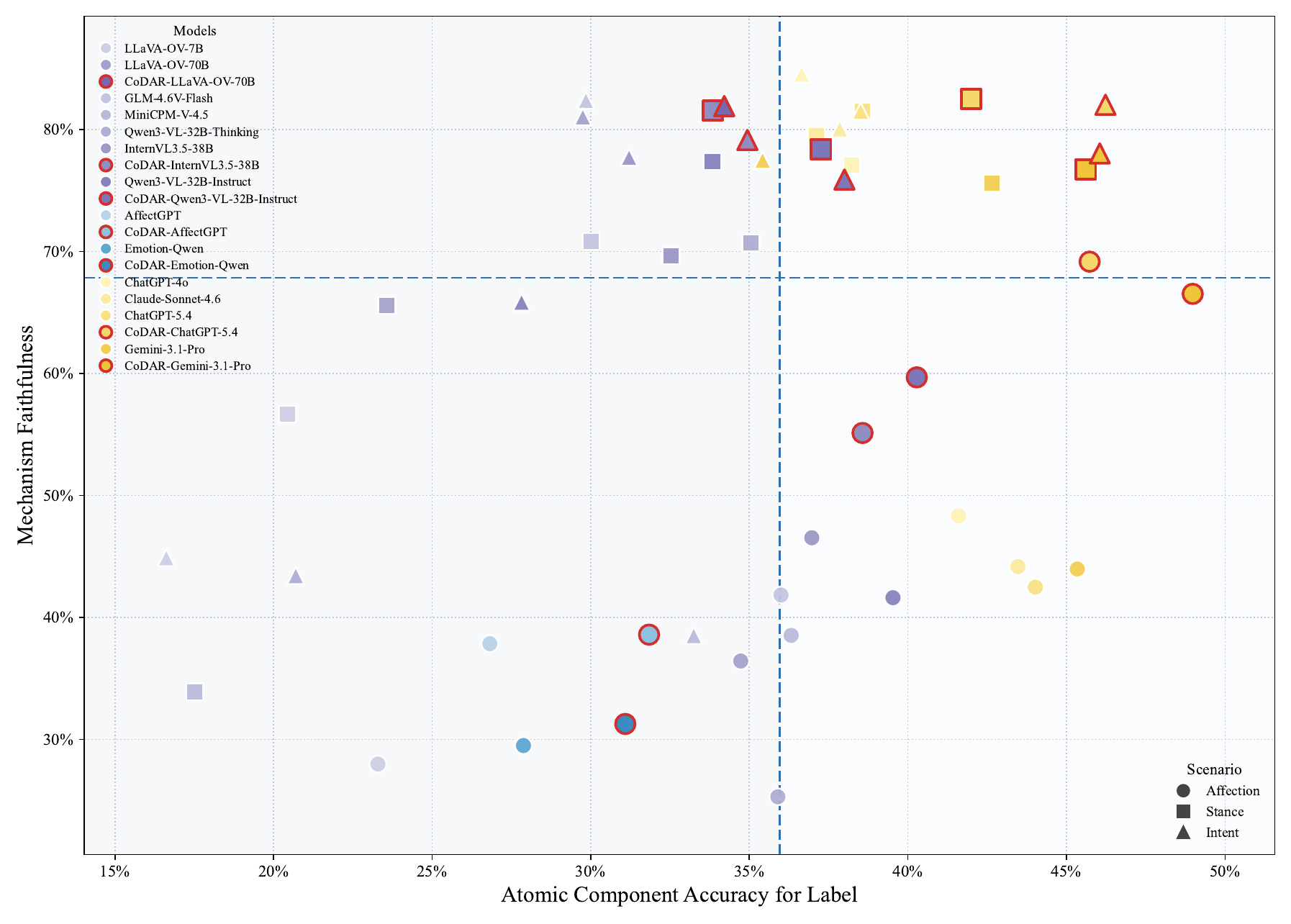}
    \caption{Scatter plot showing correlation between Atomic Component Accuracy for Label and Mechanism Faithfulness. Points with red borders denote CoDAR-augmented models, partitioned into quadrants by blue dashed lines.}
    \label{fig:label_cond_quadrant}
  \end{figure}

  \paratitle{\textit{RQ-3: Does the Thinking Variant Improve Implicit Social Reasoning?}}
  Across Tables~\ref{tab:results_affection_full}--\ref{tab:results_stance_full}, the Thinking variant records lower \textit{Mechanism} accuracy than the matched Qwen3-VL-32B-Instruct model in all scenarios. The differences are 15.92, 28.76, and 6.04 percentage points in Affection, Intent, and Stance. \textit{Holistic Alignment} is also lower by 3.84, 5.17, and 3.63 points. Target accuracy changes by $+0.53$, $+3.82$, and $+0.10$ points, whereas Subject accuracy changes by $-11.33$, $-9.58$, and $-11.17$ points. \textit{Mechanism Faithfulness} likewise changes by $-16.32$, $-22.41$, and $-6.69$ points.

  The largest differences therefore concentrate in Subject and mechanism-related metrics, whereas Target changes comparatively little. The reported results do not identify why this pattern occurs. They establish only that the Thinking variant does not improve \textit{Holistic Alignment} for this matched model pair. This conclusion should not be generalized to all thinking models or reasoning paradigms.
  
  \paratitle{\textit{RQ-4: To What Extent Do Models Achieve Label Accuracy through Mechanism-Level Understanding?}}
  A model may predict the correct label without identifying the corresponding mechanism. To quantify this discrepancy, we define a two-dimensional space using Label accuracy $x = P_L$ and Mechanism Faithfulness $y = P_{M|L}$, as shown in Figure~\ref{fig:label_cond_quadrant}. The conditional formulation measures the probability that a model predicts the correct label while failing to recover the corresponding mechanism:
  \begin{equation}
  P_{L \cap \neg M} = P_L - P_L \times P_{M|L} = P_L \left[ 1 - P_{M|L} \right].
  \end{equation}
  Higher values indicate weaker consistency between Label accuracy and mechanism recovery. Models in the upper-right region, such as ChatGPT-5.4 in Stance, achieve 38.57\% on $P_L$ and 81.46\% on $P_{M|L}$, showing consistent performance across both dimensions. In contrast, models in the lower-right region, such as Gemini-3.1-Pro in Affection, achieve comparable Label accuracy of 45.34\% but lower Mechanism Faithfulness of 43.97\%. After applying CoDAR, multiple models shift toward the upper-right region, indicating improved alignment between Label accuracy and mechanism recovery. For example, CoDAR increases Mechanism Faithfulness of ChatGPT-5.4 in Affection from 42.48\% to 69.16\%, corresponding to an improvement of 26.68\%. These shifts indicate that CoDAR better aligns label and mechanism predictions.

  \begin{table}[!t]
  \centering
  \caption{Within-family scaling from LLaVA-OneVision-7B to LLaVA-OneVision-70B. $\Delta$Dyad and $\Delta$Holis denote percentage-point changes in \textit{Dyadic Alignment} and \textit{Holistic Alignment}. $C_{\mathrm{sem}\mid\mathrm{dyad}}$ is the percentage of dyad-correct predictions that are also holistically correct.}
  \label{tab:rq5_scaling}
  \small
  \setlength{\tabcolsep}{8pt}
  {\renewcommand{\arraystretch}{1.08}
  \begin{tabular*}{\textwidth}{@{\extracolsep{\fill}}lcccc@{}}
  \toprule
  \textbf{Scenario} & $\boldsymbol{\Delta}$\textbf{Dyad} & $\boldsymbol{\Delta}$\textbf{Holis} & $\boldsymbol{C}$\textbf{: 7B} & $\boldsymbol{C}$\textbf{: 70B} \\
  \midrule
  Affection & $+37.53$ & $+4.82$ & 7.16 & 10.81 \\
  Intent    & $+35.24$ & $+13.11$ & 5.55 & 24.73 \\
  Stance    & $+39.12$ & $+7.53$ & 9.74 & 15.96 \\
  \bottomrule
  \end{tabular*}
  }
  \end{table}

  \paratitle{\textit{RQ-5: Does Within-Family Scaling Resolve Structured Implicit Reasoning?}}
  Table~\ref{tab:rq5_scaling} provides a within-family comparison between LLaVA-OneVision-7B and LLaVA-OneVision-70B. Increasing model scale raises \textit{Dyadic Alignment} by 37.53, 35.24, and 39.12 points in Affection, Intent, and Stance, respectively. The corresponding gains in \textit{Holistic Alignment} are substantially smaller at 4.82, 13.11, and 7.53 points. Thus, the larger model recovers interaction participants much more reliably, but only part of this improvement transfers to the complete four-field output.

  This separation remains visible after normalizing for dyad recovery. The conditional completion rate rises from 7.16\% to 10.81\% in Affection, from 5.55\% to 24.73\% in Intent, and from 9.74\% to 15.96\% in Stance. These values remain far below the corresponding human rates of 76.05\%, 76.41\%, and 79.80\%. The 70B checkpoint therefore alleviates, but does not remove, the compositional bottleneck. Because the two checkpoints may differ in factors beyond parameter count, this result is interpreted as a within-family contrast rather than a general scaling law.

  \FloatBarrier

  \begin{table}[!t]
  \centering
  \caption{Semantic completion after correct Subject--Target recovery. Baseline and CoDAR values are macro-averaged over InternVL3.5-38B, Qwen3-VL-32B-Instruct, ChatGPT-5.4, and Gemini-3.1-Pro. Values are percentages; $\Delta$CoDAR is the percentage-point gain.}
  \label{tab:rq6_conditional_completion}
  \small
  \setlength{\tabcolsep}{8pt}
  {\renewcommand{\arraystretch}{1.08}
  \begin{tabular*}{\textwidth}{@{\extracolsep{\fill}}lcccc@{}}
  \toprule
  \textbf{Scenario} & \textbf{Human} & \textbf{Baseline Avg.} & \textbf{CoDAR Avg.} & $\boldsymbol{\Delta}$\textbf{CoDAR} \\
  \midrule
  Affection & 76.05 & 17.78 & 21.97 & $+4.18$ \\
  Intent    & 76.41 & 24.69 & 31.73 & $+7.05$ \\
  Stance    & 79.80 & 27.66 & 31.31 & $+3.65$ \\
  \bottomrule
  \end{tabular*}
  }
  \end{table}

  \paratitle{\textit{RQ-6: What Remains Unresolved after Correct Subject--Target Recovery?}}
  Since every holistically correct prediction is necessarily dyad-correct, we define the conditional semantic completion rate as
  \begin{equation}
  C_{\mathrm{sem}\mid\mathrm{dyad}}
  =
  \frac{\textit{Holistic Alignment}}{\textit{Dyadic Alignment}}
  \times 100.
  \end{equation}
  Table~\ref{tab:rq6_conditional_completion} reports this diagnostic, which estimates how often a model also recovers the correct Mechanism and Label once both interaction participants are correct. Across the four matched backbones, the baseline macro-average is 17.78\%, 24.69\%, and 27.66\% for Affection, Intent, and Stance, respectively. CoDAR increases these rates to 21.97\%, 31.73\%, and 31.31\%, corresponding to gains of 4.18, 7.05, and 3.65 points.

  The remaining gap is nevertheless large. Even with CoDAR, fewer than one third of dyad-correct predictions become complete structured outputs in any scenario, whereas human completion ranges from 76.05\% to 79.80\%. The principal residual error therefore lies after participant identification: models still struggle to bind the recovered relation to a compatible mechanism and latent social label.

  \FloatBarrier

  \begin{figure}[!t]
  \centering
  \includegraphics[width=\textwidth]{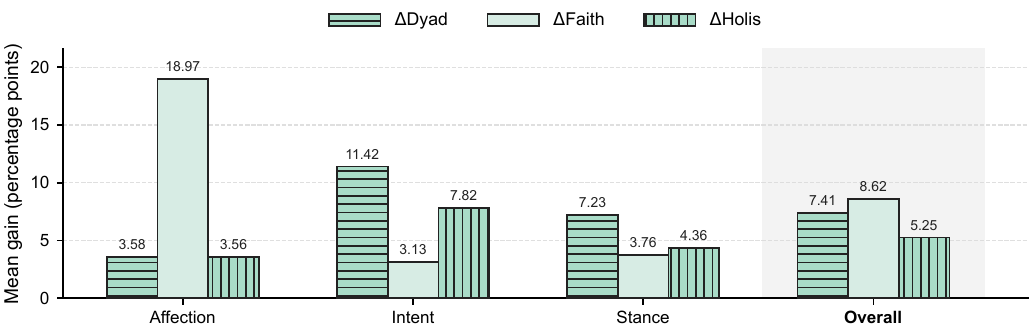}
  \caption{Scenario-wise localization of CoDAR gains. Bars report mean percentage-point improvements over InternVL3.5-38B, Qwen3-VL-32B-Instruct, ChatGPT-5.4, and Gemini-3.1-Pro. Horizontally hatched, solid, and vertically hatched bars denote $\Delta$Dyad, $\Delta$Faith, and $\Delta$Holis, respectively; numbers give exact values, and the shaded Overall group gives the macro-average across scenarios.}
  \label{fig:rq7_scenario_gains}
  \end{figure}

  \paratitle{\textit{RQ-7: Does CoDAR Improve the Same Capability in Every Scenario?}}
  We localize CoDAR's gains through 12 matched comparisons comprising four backbones evaluated before and after CoDAR in all three scenarios. As shown in Figure~\ref{fig:rq7_scenario_gains}, the gain profile is scenario-dependent. In Affection, the dominant change is in \textit{Mechanism Faithfulness} ($+18.97$ points on average), whereas \textit{Dyadic Alignment} and \textit{Holistic Alignment} rise by 3.58 and 3.56 points. In Intent, the largest gain shifts to \textit{Dyadic Alignment} ($+11.42$), accompanied by the strongest average improvement in \textit{Holistic Alignment} ($+7.82$). Stance exhibits a more balanced profile, with gains of 7.23, 3.76, and 4.36 points on the three metrics.

  All 12 model--scenario pairs improve simultaneously on \textit{Dyadic Alignment}, \textit{Mechanism Faithfulness}, and \textit{Holistic Alignment}; their overall mean gains are 7.41, 8.62, and 5.25 points, respectively. CoDAR therefore does not obtain its holistic gains by trading relation recovery against mechanism consistency. Instead, the relative source of improvement changes with the scenario, consistent with the different conflict structures that Affection, Intent, and Stance impose.

  \begin{figure}[!t]
      \centering
      \includegraphics[width=\textwidth]{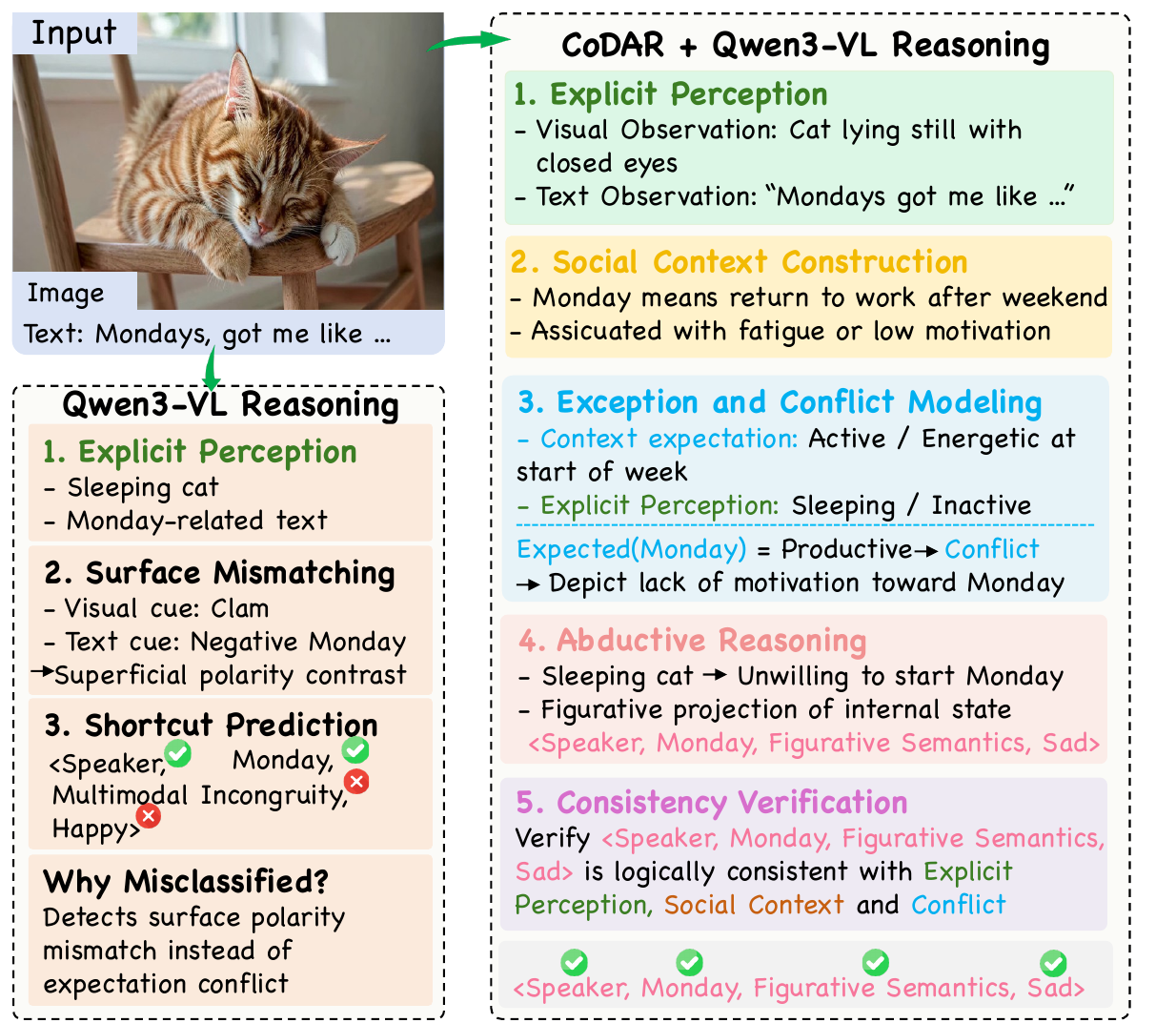} 
      \caption{Visual comparison between Qwen3-VL-32B-Instruct w/ vs. w/o our CoDAR framework, where the baseline remains limited to surface-level prediction, while CoDAR proceeds through conflict modeling and abductive reasoning.}
      \label{fig:case} 
  \end{figure}

  \subsection{Qualitative Case Study}
  \label{sec:qualitative_case}
  To illustrate the performance gap between baselines and CoDAR, we present a representative test case, as shown in Figure~\ref{fig:case}. The baseline Qwen3-VL-32B-Instruct detects polarity contrast between a calm visual cue and the negative textual cue Monday, and interprets the relation as generalized \textit{Surface Mismatching}, leading to incorrect prediction of \textit{Multimodal Incongruity} and the label Happy. In contrast, CoDAR constructs an expectation-driven reasoning process by establishing contextual expectation that Monday implies work and productivity, which conflicts with the sleeping cat. Through abductive reasoning, CoDAR interprets this conflict as figurative projection of the subject internal unwillingness to begin the week rather than simple image text inconsistency. By identifying the \textit{Figurative Semantics} mechanism and the latent Sad emotion, this example demonstrates CoDAR ability to move beyond shallow visual text matching toward implicit social reasoning grounded in relational structure.

  \section{Conclusion}
  The paper introduces a novel task, \textit{Implicit Social Context Analysis} (MoCA), that formalizes implicit social context analysis as a structured multimodal reasoning. 
  By requiring models to jointly recover the subject, target, mechanism, and latent social state across Affection, Intent, and Stance scenarios, MoCA provides the first systematic evaluation of whether multimodal large models can perform human-level cognitive reasoning in implicit social interactions. 
  We construct a high-quality benchmark with fine-grained cognitive annotations and show that current multimodal large language models struggle significantly on this task due to their reliance on explicit cues.  
  To address this limitation, we propose the \textit{Conflict-Driven Abductive Reasoning} (CoDAR) framework, which models cognitive conflict to infer hidden mental states.  
  Experiments reveal that current models exhibit a substantial and widening gap from human performance.
  Further results demonstrate that CoDAR substantially improves performance, while a notable gap to human reasoning remains.

  \section{Ethics Statement}
  
  The MoCA benchmark was constructed exclusively from publicly available multimodal sources, including existing datasets and supplementary real-world materials such as meme--text pairs and online discussions. The development of MoCA was supported by a structured human-annotation process. All annotators were graduate students at the master's level or above, including master's and doctoral students, and received formal training in the annotation protocol to ensure consistent and reliable annotation. The annotation process included independent annotation, cross-checking, disagreement arbitration, and final validation of the scenario, domain, culture, Subject, Target, Mechanism, and Label fields.

  \clearpage
  \bibliography{references}

\appendix
\setcounter{section}{0}
\renewcommand{\thesection}{\Alph{section}}
\renewcommand{\thesubsection}{\thesection.\arabic{subsection}}
\renewcommand{\thesubsubsection}{\thesubsection.\arabic{subsubsection}}
\numberwithin{equation}{section}

\section*{Appendix Overview}

\begin{itemize}
    \setlength{\itemsep}{2pt}
    \setlength{\parskip}{0pt}

    \item Appendix~\ref{sec:supp-task} defines the tasks and annotation spaces.
    \item Appendix~\ref{sec:supp-construction} details benchmark construction and statistics.
    \item Appendix~\ref{sec:supp-implementation} describes the CoDAR implementation and inference process.
    \item Appendix~\ref{sec:supp-prompts} lists the baseline and CoDAR prompt templates.
    \item Appendix~\ref{sec:supp-cases} presents representative case studies.
\end{itemize}
\section{Task Definition Supplement and Concept Boundary}
\label{sec:supp-task}

\subsection{Detailed Definitions and Boundaries of Three Implicit Social Scenarios}
\label{sec:supp-scenarios}

\paratitle{Affection.}
Implicit Affection concerns the subject's genuine emotional state when that state is not expressed directly.
The state may instead be conveyed through cross-modal conflict, figurative language, culture-dependent meaning, or deliberate emotional masking.
The inference target is what the subject feels, rather than the surface polarity of an isolated modality.

\paratitle{Intent.}
Implicit Intent concerns the interactional outcome that the subject seeks to produce.
The relevant goal may be concealed by humor, praise, aggression, avoidance, or another strategically framed action.
The inference target is what the subject aims to achieve, rather than the emotion accompanying the behavior.

\paratitle{Stance.}
Implicit Stance concerns the subject's evaluative and relational position toward a target.
It includes support or opposition, relational distance, and power posture.
The inference target is how the subject positions themself with respect to the target, rather than the subject's private emotion or desired outcome.

\paratitle{Boundary rule.}
One observation may support interpretations in all three domains.
Annotation follows the central latent variable required to explain the expression: Affection asks what is felt, Intent asks what outcome is pursued, and Stance asks what relational position is adopted.
Surface emotion words, rhetorical form, and visual salience do not determine the scenario by themselves.

\subsection{Data Field Definitions}
\label{sec:supp-fields}

Each instance is represented as
\[
(\mathrm{Scenario},\mathrm{Domain},\mathrm{Culture},
\mathrm{Subject},\mathrm{Target},\mathrm{Mechanism},\mathrm{Label}).
\]
The first three fields delimit the interpretation context, Subject and Target recover the directed social relation, and Mechanism and Label distinguish the construction of implicit meaning from its latent outcome.

\begin{table}[!t]
\centering
\caption{Operational definitions of the seven annotation fields.}
\label{tab:supp-fields}
\small
\renewcommand{\arraystretch}{1.08}
\begin{tabular*}{\textwidth}{@{\extracolsep{\fill}}p{0.16\textwidth}p{0.79\textwidth}@{}}
\toprule
\textbf{Field} & \textbf{Operational definition} \\
\midrule
Scenario & Interpretive domain: Affection, Intent, or Stance. It constrains the valid Mechanism and Label sets. \\
Domain & Social setting in which the interaction occurs: Networked, Occupational, Civic, Intimate, Familial, Peer, or Educational. \\
Culture & Background knowledge required for interpretation: General, Middle Eastern, North American, South Asian, or East Asian. \\
Subject & Social agent whose latent state, goal, or position is inferred. The agent may be visible, off-screen, collective, or contextually presupposed. \\
Target & Person, group, institution, issue, event, or environment toward which the implicit expression is directed. \\
Mechanism & Expressive or socio-cognitive strategy through which the latent meaning is constructed and concealed. \\
Label & Scenario-specific latent social state recovered from the complete multimodal evidence. \\
\bottomrule
\end{tabular*}
\end{table}

\subsection{Mechanism Space Overview}
\label{sec:supp-output-spaces}

Mechanism is treated as an explanatory bridge between observable evidence and the final Label.
The output spaces are scenario-specific so that a plausible verbal explanation cannot drift into a mechanism or label belonging to a different scenario.

\begin{table}[!t]
\centering
\caption{Mechanism inventory retained by the current MoCA formulation.}
\label{tab:supp-mechanisms}
\footnotesize
\renewcommand{\arraystretch}{1.05}
\begin{tabular*}{\textwidth}{@{\extracolsep{\fill}}p{0.12\textwidth}p{0.25\textwidth}p{0.55\textwidth}@{}}
\toprule
\textbf{Scenario} & \textbf{Mechanism} & \textbf{Decision criterion} \\
\midrule
\multirow{4}{*}{Affection}
& Multimodal Incongruity & Cross-modal semantic or polarity conflict is the decisive cue to the genuine emotion. \\
& Figurative Semantics & Metaphor, exaggeration, understatement, or another nonliteral mapping conveys the emotion. \\
& Socio-cultural Dependency & Interpretation depends on external knowledge, a cultural convention, or shared in-group context. \\
& Affective Deception & The outward display masks, suppresses, or substitutes for the genuine emotion. \\
\midrule
\multirow{4}{*}{Intent}
& Expressive Aggression & Aggressive expression is used strategically to steer the interaction. \\
& Malicious Manipulation & Apparently reasonable behavior conceals a self-serving or harmful objective. \\
& Prosocial Deception & Concealment or apparent agreement protects another person or preserves the relationship. \\
& Benevolent Provocation & Provocation or misdirection is used to elicit an action intended to benefit the target. \\
\midrule
\multirow{4}{*}{Stance}
& Dominant Detachment & A superior position is combined with distancing, exclusion, or contempt. \\
& Dominant Affiliation & Apparent support or acceptance preserves a top-down relational asymmetry. \\
& Protective Distancing & Withdrawal or reduced commitment protects the subject under risk or unequal power. \\
& Submissive Alignment & Accommodation or excessive agreement lowers the subject's position to preserve security or attachment. \\
\bottomrule
\end{tabular*}
\end{table}
\FloatBarrier

\subsection{Label Space Overview}

The label inventory encodes the scenario-specific latent social outcome and is interpreted jointly with the corresponding Mechanism rather than as a free-form description.
Table~\ref{tab:supp-labels} lists the admissible labels used for structured evaluation.

\begin{table}[!t]
\centering
\caption{Valid label sets under each scenario.}
\label{tab:supp-labels}
\small
\renewcommand{\arraystretch}{1.08}
\begin{tabular*}{\textwidth}{@{\extracolsep{\fill}}p{0.15\textwidth}p{0.80\textwidth}@{}}
\toprule
\textbf{Scenario} & \textbf{Labels} \\
\midrule
Affection & Happy, Sad, Disgusted, Angry, Fearful, Bad. \\
Intent & Mitigate, Intimidate, Alienate, Mock, Denounce, Provoke, Dominate, Condemn. \\
Stance & Supportive, Appreciative, Sympathetic, Neutral, Indifferent, Disapproving, Skeptical, Concerned, Dismissive, Contemptuous, Hostile. \\
\bottomrule
\end{tabular*}
\end{table}

\section{MoCA Construction and Annotation Details}
\label{sec:supp-construction}

\subsection{Construction Pipeline}
\label{sec:supp-sources}

The source pool combines established multimodal benchmarks with publicly accessible real-world material.
Video-text candidates are drawn from MUStARD~\citep{castro2019towards} and MIntRec~\citep{zhang2022mintrec}.
Image-text candidates include RedCaps~\citep{desai2021redcaps}, MMSD2.0~\citep{qin2023mmsd2}, Hateful Memes~\citep{kiela2020hatefulmemes}, MultiMET~\citep{zhang2021multimet}, MemeCap~\citep{hwang2023memecap}, Instagram captions, meme-text pairs, and online debates.
These sources provide complementary evidence from conversational video, social-media imagery, figurative expression, semantic incongruity, and public interaction.

Preprocessing removes invalid links, markup, system artifacts, corrupted media, and duplicate candidates while preserving semantically relevant emojis and hashtags.
Video frames remain temporally ordered, and speech is represented by transcripts that preserve linguistic content and speaking style.
Consequently, the reported audio-related setting evaluates transcript-conditioned multimodal reasoning rather than direct acoustic perception.

\subsubsection{Unimodal Consistency Filtering}
\label{sec:supp-filtering}

Candidate instances are evaluated under visual-only, text-only, and multimodal conditions.
Let \(R_{\mathrm{vis}}\) denote the image-only or video-only interpretation, \(R_{\mathrm{txt}}\) the text-only interpretation, and \(R_{\mathrm{multi}}\) the interpretation from the complete input.
A candidate proceeds to structured annotation only when
\begin{equation}
R_{\mathrm{multi}}\neq R_{\mathrm{vis}}
\quad\text{and}\quad
R_{\mathrm{multi}}\neq R_{\mathrm{txt}}.
\label{eq:supp-filter}
\end{equation}
The criterion rejects cases for which a single modality already determines the intended interpretation.
It is a screening rule for cross-modal dependence, not a claim that every retained instance contains a literal polarity reversal.

\subsection{Annotation}
\label{sec:supp-annotation}

All annotators were graduate students at the master's level or above, including master's and doctoral students.
They received formal training on scenario boundaries, field definitions, mechanism criteria, and common sources of multimodal ambiguity.
The operational workflow comprised four stages:
\begin{compactenum}
\item \textbf{Independent annotation.} An annotator reviewed the complete multimodal input and assigned all seven fields under a shared codebook.
\item \textbf{Field-level cross-checking.} A separate reviewer examined scenario assignment, Subject--Target direction, Mechanism--Label compatibility, and evidential support.
\item \textbf{Disagreement arbitration.} Disputed fields were revised, retained, or rejected after joint inspection of the multimodal evidence and annotation rationale.
\item \textbf{Final validation.} Every retained sample was checked for a valid scenario, a coherent directed relation, a scenario-valid Mechanism and Label, and sufficient multimodal support.
\end{compactenum}

\subsection{Quality Control}
\label{sec:supp-quality-control}

This process treats disagreement as structural rather than label-only.
For example, matching labels do not constitute agreement when annotators assign different Subjects, reverse the relation direction, or select a Mechanism that does not explain the Label.
Numerical agreement coefficients are not introduced here because the current released materials do not include the per-annotator records required to reproduce them.

\subsection{Detailed Statistics}
\label{sec:supp-composition}

The final benchmark contains 3,108 instances.
Its 936 Affection, 1,179 Intent, and 993 Stance samples correspond to 30.12\%, 37.93\%, and 31.95\% of the benchmark.
The modality split comprises 2,230 image-text and 878 video-text instances.
All seven structured fields are populated, yielding 21,756 final field values.

\begin{table}[!t]
\centering
\caption{Structural and modality composition of the final MoCA benchmark.}
\label{tab:supp-composition}
\small
\renewcommand{\arraystretch}{1.08}
\begin{tabular*}{\textwidth}{@{\extracolsep{\fill}}p{0.68\textwidth}rr@{}}
\toprule
\textbf{Partition} & \textbf{Count} & \textbf{Share} \\
\midrule
Affection & 936 & 30.12\% \\
Intent & 1,179 & 37.93\% \\
Stance & 993 & 31.95\% \\
\midrule
Image-text & 2,230 & 71.75\% \\
Video-text & 878 & 28.25\% \\
\midrule
Final instances & 3,108 & 100.00\% \\
Populated seven-field values & 21,756 & 100.00\% \\
\bottomrule
\end{tabular*}
\end{table}

\section{CoDAR Details}
\label{sec:supp-implementation}

\subsection{Evaluation and Input Interface}
\label{sec:supp-evaluation}

Models and humans receive the same scenario-specific instances and return the same four evaluated fields:
\((\mathrm{Subject},\mathrm{Target},\mathrm{Mechanism},\mathrm{Label})\).
Predictions and annotations are lowercased, hyphens and underscores are mapped to spaces, and repeated separators are collapsed before exact matching.
A missing, unparsable, or scenario-invalid field is counted as incorrect; the corresponding sample is not removed.
This implementation supports the four metrics defined in the main paper without selective filtering.

Image-text inputs contain the original text and paired image.
Video-text inputs contain the original text and a temporally ordered sequence of sampled frames.
Speech is supplied as a transcript and processed jointly with the visual stream.
No result in the current manuscript should therefore be interpreted as evaluation of uninterrupted raw-video or direct acoustic perception.

\subsection{Details of Architecture}
\label{sec:supp-stage-contract}

CoDAR is implemented as a train-free sequence of instruction-guided stages.
The stages pass structured intermediate representations rather than optimize a learned scoring function or a separate decoding objective.
Table~\ref{tab:supp-stage-contract} records the minimum contract used to keep this process reproducible.

\begin{table}[!t]
\centering
\caption{Stage-level input, inference objective, and output contract of CoDAR.}
\label{tab:supp-stage-contract}
\small
\setlength{\tabcolsep}{4pt}
\renewcommand{\arraystretch}{1.06}
\begin{tabular*}{\textwidth}{@{\extracolsep{\fill}}c
p{0.17\textwidth}
p{0.43\textwidth}
p{0.24\textwidth}@{}}
\toprule
\textbf{Stage} & \textbf{Input} & \textbf{Required inference} & \textbf{Output} \\
\midrule
1 & Multimodal observation & Record only observable linguistic, visual, temporal, and transcript evidence. & Explicit perception \(x\). \\
2 & \(x\), internal knowledge, optional retrieval & Identify missing social background and synthesize relevant facts, relations, and norms. & Social context \(c\). \\
3 & \(x,c\) & Derive the contextually expected expression \(e\), then locate the contradiction between \(e\) and \(x\). & Expectation \(e\) and conflict \(\mathcal C\). \\
4 & \(x,c,e,\mathcal C\), valid candidates & Analyze counterfactual cost and the strategic function of conflict; recover one admissible structured interpretation. & Candidate tuple \(\mathcal A\). \\
5 & \(\mathcal A,x,c,\mathcal C\) & Check evidence, relation, scenario, and mechanism--label consistency; revise only the failed component chain. & Verified tuple or targeted revision. \\
\bottomrule
\end{tabular*}
\end{table}

\subsection{Conflict-Constrained Staged Inference}
\label{sec:supp-constrained}

For instance \(i\) in scenario \(s\), let \(O_{\mathrm{subj}}^{(i)}\) and \(O_{\mathrm{tgt}}^{(i)}\) denote its Subject and Target candidate sets.
Let \(\mathcal M_s\) and \(\mathcal L_s\) denote the scenario-valid Mechanism and Label sets.
The admissible output space is
\begin{equation}
\Omega_{i,s}
=
O_{\mathrm{subj}}^{(i)}
\times O_{\mathrm{tgt}}^{(i)}
\times \mathcal M_s
\times \mathcal L_s .
\label{eq:supp-space}
\end{equation}
The model must return exactly one element from each component set:
\begin{equation}
\mathcal A_i
=
\{y_{\mathrm{subj}},y_{\mathrm{tgt}},
y_{\mathrm{mech}}^{s},y_{\mathrm{label}}^{s}\}
\in\Omega_{i,s}.
\label{eq:supp-tuple}
\end{equation}

Inference proceeds in a fixed order.
Counterfactual cost asks what social cost would arise if the subject expressed the latent state directly.
Conflict utility asks how the observed deviation reduces that cost or advances an interactional goal.
Participant recovery identifies who deploys the strategy and toward whom it is directed.
Mechanism--Label assignment then selects the scenario-valid explanatory bridge and latent outcome jointly.
This order constrains open-ended verbal reasoning without claiming an explicit combinatorial search.

The final stage evaluates
\begin{equation}
\Phi(\mathcal A_i,x_i,c_i,\mathcal C_i)\in\{0,1\}.
\label{eq:supp-verification}
\end{equation}
A value of zero triggers a targeted return to the stage responsible for the failed evidence, context, relation, or mechanism--label link.
A value of one indicates output-level consistency only; it does not establish that the model's internal reasoning is causally faithful.

Operationally, a failed check is localized through the stage dependency chain.
Unsupported observable evidence returns to Stage~1; an insufficient social premise returns to Stage~2; an unresolved expectation--observation conflict returns to Stage~3; and a scenario-invalid or internally incompatible tuple returns to Stage~4.
Components that remain supported are retained, so revision targets the failed dependency rather than restarting the complete inference chain.

\section{Prompt Templates}
\label{sec:supp-prompts}

Because CoDAR is training-free, its prompts are the method rather than an
implementation detail, so we state them explicitly.
Prompt~1 is the direct-inference prompt used for baseline models
(Appendix~\ref{sec:supp-prompt-baseline}); Prompts~2--7 are the stage prompts
that instantiate CoDAR (Appendix~\ref{sec:supp-prompt-codar}).

Two conventions apply throughout.
First, \texttt{\{...\}} marks a slot filled at runtime, following the notation of
the released templates; the inventories injected into \texttt{\{mechanisms\}} and
\texttt{\{labels\}} are those of Table~\ref{tab:supp-mechanisms} and
Table~\ref{tab:supp-labels}.
Second, the boxes below are \emph{abridged}: they retain each prompt's role,
inputs, operative instructions, constraints, and output contract, and drop only
output-formatting boilerplate and the per-option boundary rules.
The unabridged templates are released verbatim with the code, in
\texttt{moca/baseline\_prompts.py} (Prompt~1) and \texttt{moca/prompts.py}
(Prompts~2--7).

\subsection{Direct-Inference Baseline Prompt}
\label{sec:supp-prompt-baseline}

Baseline models receive one scenario-conditioned system prompt and emit the four
fields in a single pass, so a model is never asked to choose among values
belonging to another scenario.
Subject and Target candidates appear as shuffled slot identifiers under a
per-instance deterministic seed, and the model returns an identifier rather than
a surface string; this removes position and verbatim-copy shortcuts and makes
exact matching unambiguous.
Decoding uses temperature~0 with JSON-mode output, and a missing, unparsable, or
scenario-invalid field is scored as incorrect
(Appendix~\ref{sec:supp-evaluation}).

\begin{promptbox}{Direct structured prediction (baseline)}
\textbf{[Task]} Objectively classify the implicit social dynamics in the provided text--image or text--video pair. This is a strict technical annotation task for sociological research.\par
\textbf{[Terminology]} The category names denote abstract constructs in pragmatics. They are academic codes for communicative strategies, not moral judgments about any individual depicted.\par
\textbf{[Input]} \texttt{\{Text\}}, \texttt{\{Visual\}}, \texttt{\{Audio caption\}}, and the slot mappings \texttt{subject0..3}, \texttt{target0..3}.\par
\textbf{[Output space]} Exactly four fields, one value each: \texttt{mechanism} from \texttt{\{mechanisms\}}; \texttt{label} from \texttt{\{labels\}}; \texttt{subject} from \texttt{subject0..3}; \texttt{target} from \texttt{target0..3}. Every option is supplied with its decision criterion and boundary rule.\par
\textbf{[Rules]} Return only a strict JSON object with exactly these four keys. For \texttt{subject} and \texttt{target}, output the slot identifier, never the descriptive text. Do not include explanations, rationales, or chain-of-thought reasoning.\par
\textbf{[Output]} \texttt{\{"mechanism":..., "label":..., "subject":..., "target":...\}}
\end{promptbox}

\subsection{CoDAR Stage Prompts}
\label{sec:supp-prompt-codar}

CoDAR replaces that single call with six role-conditioned prompts over the five
stages of Table~\ref{tab:supp-stage-contract}, Stage~2 being split into query
formulation and context synthesis.
Each prompt states what its stage must produce and, equally, what it must not yet
infer; these negative constraints are what keep an early stage from collapsing
into a direct prediction.
The two families share the evaluated output space, the slot-identifier protocol,
and greedy decoding, and differ in whether the four fields are produced directly
or through five staged intermediate representations.

\begin{promptbox}{Stage 1 --- Observer}
\textbf{[Role]} Act as a forensic camera and microphone.\par
\textbf{[Input]} \texttt{\{Text\}}, \texttt{\{Visual\}}, \texttt{\{Audio\}}\par
\textbf{[Task]} Produce one two-to-four sentence caption recording only directly observable entities, spoken words, on-screen text, physical actions or states, and spatial relations. For video, follow chronological order.\par
\textbf{[Forbidden]} Metaphor or intent decoding; sociological, psychological, or functional interpretation; emotion or stance labels.\par
\textbf{[Output]} \texttt{\{Explicit Perception\}}
\end{promptbox}

\begin{promptbox}{Stage 2a --- Query Formulator}
\textbf{[Role]} Act as a neutral background investigator.\par
\textbf{[Input]} \texttt{\{Explicit Perception\}}\par
\textbf{[Task]} Identify entities, symbols, or text formats whose historical context, power dynamics, or cultural norms are not immediately obvious. Assign each a retrieval route (\texttt{model\_prior}, \texttt{web\_search}, or \texttt{hybrid}) and pose two to three objective retrieval questions.\par
\textbf{[Constraints]} Questions must be neutral and must not be answered at this stage. Do not ask leading questions about hidden intent or stance.\par
\textbf{[Output]} \texttt{\{Social Query\}}
\end{promptbox}

\begin{promptbox}{Stage 2b --- Context Synthesizer}
\textbf{[Role]} Act as an encyclopedic compiler.\par
\textbf{[Input]} \texttt{\{Social Query\}}, \texttt{\{Explicit Perception\}}\par
\textbf{[Task]} Recover how the selected entities operate in the real world, not what this particular author is doing, and compress the evidence into exactly three fields: \texttt{fact}, \texttt{connection}, \texttt{social\_norm}.\par
\textbf{[Constraints]} Use neutral academic phrasing, prefer externally sourced evidence over unstated assumptions, and make uncertainty explicit. Do not interpret the hidden meaning of the instance.\par
\textbf{[Output]} \texttt{\{Social Context\}}
\end{promptbox}

\begin{promptbox}{Stage 3 --- Expectation and Conflict Analyst}
\textbf{[Role]} Identify the socially meaningful deviation between explicit evidence and contextual expectation.\par
\textbf{[Input]} \texttt{\{Explicit Perception\}}, \texttt{\{Social Context\}}\par
\textbf{[Task]} Produce exactly four sections.
\texttt{Context\_Expectation}: what the context predicts, grounded in \texttt{fact}, \texttt{connection}, and \texttt{social\_norm}.
\texttt{Explicit\_Reality}: the observable surface presentation only.
\texttt{The\_Conflict}: which explicit cue departs from which expectation.
\texttt{Abductive\_Question}: why this reality is presented given that expectation.\par
\textbf{[Constraints]} Treat the conflict as a deviation requiring explanation, not as the final meaning. Do not infer the final stance, mechanism, or label.\par
\textbf{[Output]} \texttt{\{Context\_Expectation\}}, \texttt{\{Explicit\_Reality\}}, \texttt{\{The\_Conflict\}}, \texttt{\{Abductive\_Question\}}
\end{promptbox}

\begin{promptbox}{Stage 4 --- Abductive Decoder}
\textbf{[Role]} Answer the \texttt{Abductive\_Question} by explaining why the actor produced the conflict instead of a direct, expectation-aligned expression.\par
\textbf{[Input]} The four Stage-3 sections, \texttt{\{mechanisms\}}, \texttt{\{labels\}}, \texttt{\{options.subject\}}, \texttt{\{options.target\}}\par
\textbf{[Task]} Complete four substeps in order.
(i) \emph{Counterfactual cost}: assume direct expectation-aligned expression and identify its likely social, relational, strategic, or power cost.
(ii) \emph{Conflict utility}: treat the explicit reality as a strategic mask and explain how it avoids that cost while still advancing the hidden goal.
(iii) \emph{Subject--Target recovery}: identify who deploys the strategy and toward whom, using communicative structure rather than the most salient entity.
(iv) \emph{Mechanism--Label anchoring}: select one mechanism and one label consistent with substeps (i)--(iii).\par
\textbf{[Constraints]} Return exactly one value per field, each drawn from the supplied candidate set. Do not invent new options or produce free-form essays.\par
\textbf{[Output]} \texttt{\{Subject\}}, \texttt{\{Target\}}, \texttt{\{Mechanism\}}, \texttt{\{Label\}}
\end{promptbox}

\begin{promptbox}{Stage 5 --- Consistency Checker}
\textbf{[Role]} Reverse-check whether the decoded tuple forms a logically closed chain.\par
\textbf{[Input]} \texttt{\{Explicit Perception\}}, \texttt{\{Social Context\}}, \texttt{\{The\_Conflict\}}, \texttt{\{Subject\}}, \texttt{\{Target\}}, \texttt{\{Mechanism\}}, \texttt{\{Label\}}\par
\textbf{[Task]} Evaluate three dimensions.
\emph{Explicit evidence}: supported by observable cues only, rejecting speculation.
\emph{Social context}: compatible with \texttt{fact}, \texttt{connection}, and \texttt{social\_norm}, rejecting fabricated background.
\emph{Mechanism utility}: the mechanism explains both the conflict and its indirect utility, so that mechanism, conflict, and latent meaning form a coherent chain.\par
\textbf{[Verdict]} If all checks pass, return \texttt{ACCEPTED}. Otherwise return \texttt{REJECTED} and identify the breakpoint.\par
\textbf{[Constraints]} Verify the existing interpretation rather than generating a new one. Do not introduce new evidence, mechanism, label, subject, or target.\par
\textbf{[Output]} \texttt{\{Status\}}, \texttt{\{Refine\}}
\end{promptbox}

\section{Case Study}
\label{sec:supp-cases}

The main paper provides one direct-versus-CoDAR comparison.
The two cases below illustrate complementary failure modes; they are
descriptive and do not replace the aggregate evaluation.

\begin{figure}[!t]
    \centering
    \includegraphics[width=\columnwidth]{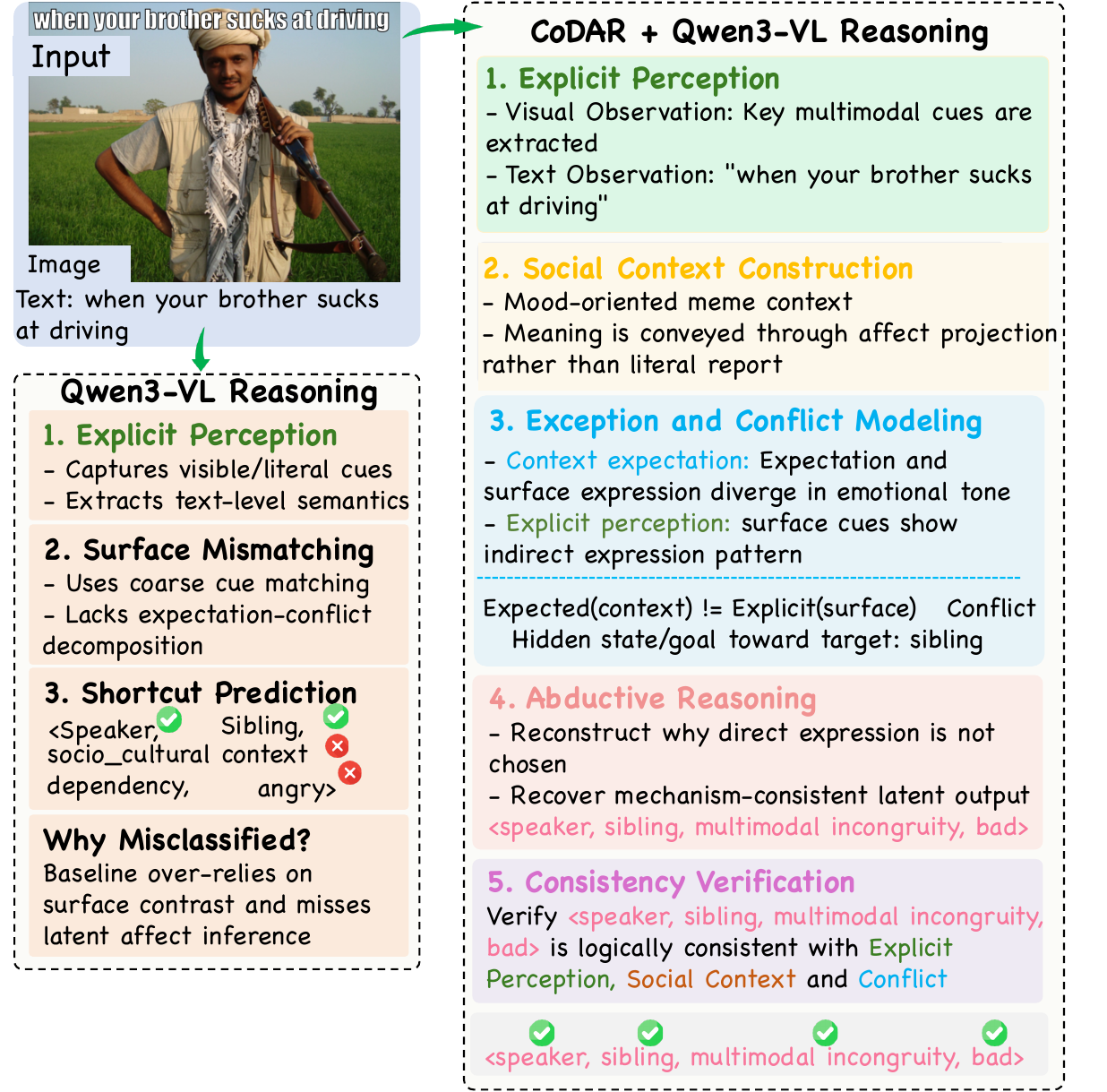}
    \caption{A case in the Affection scenario.}
    \label{fig:supp-affection-case}
\end{figure}

\noindent\textbf{Affection case.}
In Figure~\ref{fig:supp-affection-case}, correct participant recovery is
insufficient for a correct structured interpretation.
The baseline recognizes the participants but assigns a mechanism that
does not explain the latent affect.
CoDAR instead anchors the conflict to the Affection-specific output space.

\FloatBarrier

\begin{figure}[!t]
    \centering
    \includegraphics[width=\columnwidth,height=0.8\textheight,keepaspectratio]{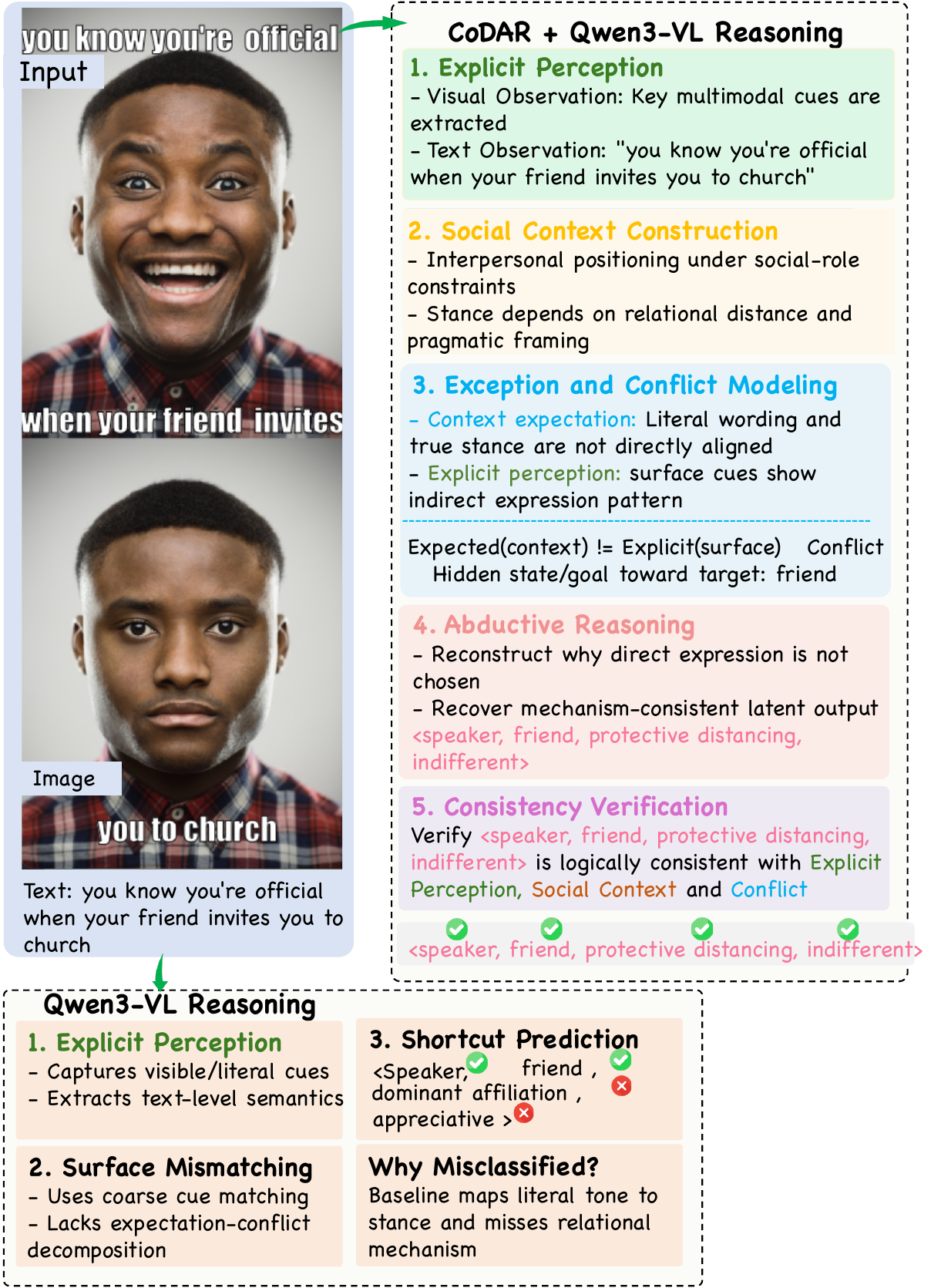}
    \caption{A case in the Stance scenario.}
    \label{fig:supp-stance-case}
\end{figure}

\noindent\textbf{Stance case.}
Figure~\ref{fig:supp-stance-case} separates positive surface presentation
from relational positioning.
The visible smile and affirmative wording do not establish affiliation
by themselves.
CoDAR instead interprets the expression as protective withdrawal under
a social-role constraint.

\newpage

\noindent\textbf{Cross-case interpretation.}
Across both examples, the baseline identifies surface cues but maps them
directly to an output.
CoDAR instead requires the selected Mechanism to explain both the observed
conflict and the recovered Subject--Target relation.
Accordingly, \(\Omega_{i,s}\) removes scenario-invalid components, whereas
\(\Phi\) checks whether the remaining tuple is jointly supported.
The cases illustrate output-level revision rather than internal causal
faithfulness.

\end{document}